\documentclass[10pt,journal,compsoc]{IEEEtran}
\ifCLASSOPTIONcompsoc
  \usepackage[nocompress]{cite}
\else
  \usepackage{cite}
\fi

\ifCLASSINFOpdf
\else
\fi

\usepackage{amsmath}
\usepackage{amssymb}
\usepackage{graphicx}
\usepackage{hyperref}
\usepackage{color}
\usepackage{multirow}
\usepackage{subfigure}
\usepackage{enumitem}
\newtheorem{proposition}{Proposition}
\newcommand{\tabincell}[2]{\begin{tabular}{@{}#1@{}}#2\end{tabular}}  

\begin{document}
%
\title{Deep CNNs Meet Global Covariance Pooling: Better Representation and Generalization}
%
%
%
%

\author{Qilong~Wang,~\IEEEmembership{Member,~IEEE,}
	    Jiangtao~Xie,
	    Wangmeng~Zuo,~\IEEEmembership{Senior Member,~IEEE,}
	    Lei~Zhang,~\IEEEmembership{Fellow,~IEEE,}
        and~Peihua~Li,~\IEEEmembership{Member,~IEEE}
\IEEEcompsocitemizethanks{\IEEEcompsocthanksitem Q. Wang is with Tianjin Key Lab of Machine Learning, the College of Intelligence and Computing, Tianjin University, Tianjin 300350, China and with the School of Information and Communication Engineering, Dalian University of Technology, Liaoning 116024, China. (E-mail: qlwang@tju.edu.cn)
\IEEEcompsocthanksitem J. Xie and P. Li are with the School of Information and Communication Engineering, Dalian University of Technology, Liaoning 116024, China. (E-mail: jiangtaoxie@mail.dlut.edu.cn; peihuali@dlut.edu.cn)
\IEEEcompsocthanksitem W. Zuo is with the School of Computer Science and
Technology, Harbin Institute of Technology, Harbin, 150001, China. (E-mail: cswmzuo@gmail.com)
\IEEEcompsocthanksitem L. Zhang is with the Department of Computing, The Hong Kong Polytechnic University, Hung Hom, Hong Kong. (E-mail: cslzhang@comp.polyu.edu.hk)
\IEEEcompsocthanksitem Peihua Li is the corresponding author. 
(E-mail: peihuali@dlut.edu.cn)}}

%
%

\markboth{Accepted to IEEE Transactions on Pattern Analysis and Machine Intelligence}%
{Shell \MakeLowercase{\textit{et al.}}: Bare Demo of IEEEtran.cls for Computer Society Journals}
%



\IEEEtitleabstractindextext{%
\begin{abstract}
Compared with global average pooling in existing deep convolutional neural networks (CNNs), global covariance pooling can capture richer statistics of deep features, having potential for improving representation and generalization abilities of deep CNNs. However, integration of global covariance pooling into deep CNNs brings two challenges: (1) robust covariance estimation given deep features of high dimension and small sample size; (2) appropriate usage of geometry of covariances. To address these challenges, we propose a \emph{global Matrix Power Normalized COVariance (MPN-COV) Pooling}. Our MPN-COV conforms to a robust covariance estimator, very suitable for scenario of high dimension and small sample size. It can also be regarded as Power-Euclidean metric between covariances, effectively exploiting their geometry. Furthermore, a global Gaussian embedding network is proposed to incorporate first-order statistics into MPN-COV. For fast training of MPN-COV networks, we implement an iterative matrix square root normalization, avoiding GPU unfriendly eigen-decomposition inherent in MPN-COV. Additionally, progressive $1\times1$ convolutions and group convolution are introduced to compress covariance representations. The proposed methods are highly modular, readily plugged into existing deep CNNs. Extensive experiments are conducted on large-scale object classification, scene categorization, fine-grained visual recognition and texture classification, showing our methods outperform the counterparts and obtain state-of-the-art performance.
\end{abstract}

}

\maketitle

\IEEEdisplaynontitleabstractindextext

%
\IEEEpeerreviewmaketitle

\IEEEraisesectionheading{\section{Introduction}\label{sec:introduction}}


%
%
%
%
\IEEEPARstart{D}{eep} learning methods,  particularly deep convolutional neural networks (CNNs),  have achieved great success in many computer vision tasks, especially in visual recognition  \cite{nips2012cnn,Simonyan15,Szegedy_2015_CVPR,He_2016_CVPR,Huang_2017_CVPR}. As described in~\cite{Bengio-PAMI-RL}, deep learning automatically learns composition of multiple  transformations that aims to produce abstract representations for predictions, and a good representation is  one that can capture distribution of input data and is discriminative as an input of supervised prediction. However, existing deep CNNs often use global average pooling (GAP) to summarize the last convolution features as input of prediction, which captures only first-order statistics, limiting representation and generalization abilities of deep CNNs. To overcome this problem, one possible solution is to replace first-order GAP by some more powerful statistical modeling methods.
 
Compared with first-order pooling methods, covariance (second-order) pooling is able to capture richer statistics of features so that it can generate more informative representations~\cite{DBLP:conf/eccv/TuzelPM06,carreira_pami14,WangLZZ16}. In the classical, shallow architectures, Tuzel et al.~\cite{DBLP:conf/eccv/TuzelPM06} for the first time  utilize covariance matrices for representing regular regions of images. Carreira et al.~\cite{carreira_pami14} present an O$_{2}$P method, where second-order non-central moments are used to model free-form regions of images. Since covariance matrices are symmetric positive definite (SPD) matrices whose space forms a Riemannian manifold, geometric structure should be favorably considered for realizing the full potential. Therefore, affine invariant Riemannian metric (AIRM) \cite{Pennec2006} and Log-Euclidean Riemannian metric (LERM) \cite{Arsigny2005} are respectively employed in~\cite{DBLP:conf/eccv/TuzelPM06}  and \cite{carreira_pami14}  to measure the distances between covariance matrices. In particular, many researches ~\cite{sanchez,carreira_pami14,WangLZZ16,KoniuszYGM17} show that covariance representations with hand-crafted features (e.g., SIFT \cite{Lowe04}) significantly outperform first-order counterparts.

In deep architectures, Ionescu et al. propose DeepO$_{2}$P networks~\cite{Ionescu_2015_ICCV} in which  second-order pooling is inserted after the last convolution layer of  deep CNNs. They establish  theory and practice of backpropagation, which enables end-to-end learning of CNNs involving non-linear structural matrices. A parallel work is bilinear CNN (B-CNN) \cite{lin2015bilinear} model, which performs outer product pooling of output features of the last convolution layers from two CNNs, producing non-central moments when the two CNNs are identical. Their main difference lies in normalization on SPD matrices obtained by covariance pooling: DeepO$_{2}$P uses matrix logarithm while B-CNN performs element-wise power normalization followed by $\ell_{2}$-normalization.

\begin{figure*}
  \centering
  \includegraphics[width=0.72\linewidth]{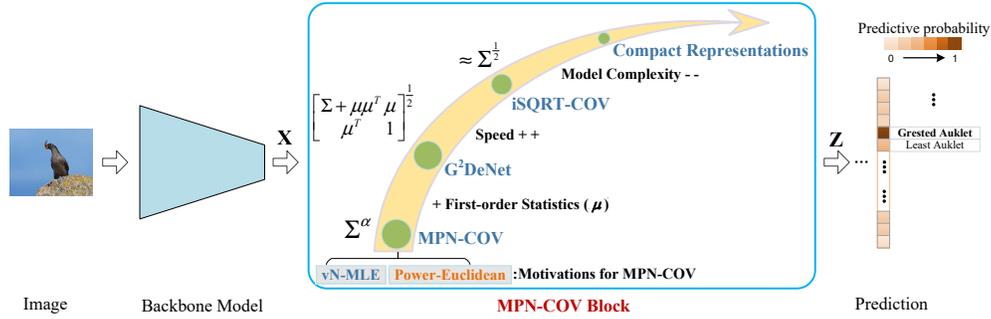}
  \caption{Overview of global Matrix Power Normalized COVariance (MPN-COV) Pooling networks. The MPN-COV block, inspired by vN-MLE~\cite{Wang_2016_CVPR} for robust covariance estimation and Power-Euclidean metric~\cite{Dryden2009} for usage of geometry of covariance matrices, is inserted after the last convolution layer of a backbone model for summarizing the second-order statistics as inputs of classifier. We further refine MPN-COV from three aspects: firstly, in G$^{2}$DeNet, we combine first-order statistics with covariance matrix by using a Gaussian, identified as matrix square root of an SPD matrix, for further performance improvement; then, an iterative matrix square root normalization (iSQRT-COV) method is developed for fast network training (Speed + +); finally, compact covariance representations are presented to reduce model complexity (Model Complexity - -).}\label{overview}
\end{figure*}

Although above global covariance pooling (GCP) methods \cite{Ionescu_2015_ICCV,lin2015bilinear} have been studied in deep architectures and report improvement, there exist two challenges that remain unresolved. The first one is robust covariance estimation. The dimensions of output features of the last convolution layers in deep CNNs usually are very high (e.g., 256 in AlexNet \cite{nips2012cnn}, 512 in VGG-VD \cite{Simonyan15} and 2048 in ResNet \cite{He_2016_CVPR}), but the feature number is small ($< 200$). Under such a scenario of high dimension and small sample size (HDSS), it is well-known that the sample covariance obtained by the classical maximum likelihood estimation (MLE) is not robust \cite{Stein2001,Donoho2014}. Furthermore, recent studies \cite{WangLZZ16,Wang_2016_CVPR} have shown that robust covariance estimation leads to clear performance improvement in the case of HDSS. The second challenge is how to make favorable use of Riemannian geometry of covariance matrices. DeepO$_{2}$P \cite{Ionescu_2015_ICCV} exploits geometry of covariances using LERM \cite{Arsigny2005}, but our experiments (Section~\ref{Comparisons}) show that LERM brings side effect for covariance matrices in the HDSS scenario. B-CNN \cite{lin2015bilinear} simply regards the space of covariances as a Euclidean space, discarding its geometric structure. Furthermore, neither DeepO$_{2}$P \cite{Ionescu_2015_ICCV} nor B-CNN \cite{lin2015bilinear} is concerned with robust covariance estimation. 

Therefore, a question naturally arises: \emph{Can we overcome above challenges that GCP faces to further improve the representation and generalization abilities of deep CNNs?} Inspired by a regularized maximum likelihood estimator (namely vN-MLE) of our previous work~\cite{Wang_2016_CVPR} and Power-Euclidean (Power-E) metric~\cite{Dryden2009}, this paper proposes a global Matrix Power Normalized COVariance (MPN-COV) Pooling to address the above two challenges. As shown in~\cite{Wang_2016_CVPR}, vN-MLE performs robustly in the HDSS scenario, superior to most of the robust covariance estimators. The solution of vN-MLE corresponds to shrinkage of eigenvalues of covariance matrices~\cite{Stein1986,ledoit2004wellconditioned}, and, interestingly, one special  solution of vN-MLE is matrix square root (power of 1/2) of sample covariance. On the other hand, Power-E metric~\cite{Dryden2009}, which computes matrix power of SPD matrices for measuring distances, has been successfully used in medical imaging. Power-E metric has close connection with LERM~\cite{Arsigny2005} and so can approximately measure the Riemannian distance on the space of covariances. As such, our matrix power normalization in the context of deep CNNs amounts to robust covariance estimation while approximately yet effectively exploiting Riemannian geometry of covariances. The statistical and geometrical mechanisms underlying MPN-COV and deep CNNs with MPN-COV (namely MPN-COV-Net) are presented in Section~\ref{subsection:MPN-COV layer} and Section~\ref{subsection: fp and bp}, respectively.

Although MPN-COV can handle aforementioned challenges, it has several downsides. Firstly, it discards first-order statistics (mean vector) that is widely used in conventional deep CNNs. In order to integrate first-order information, we use a Gaussian model that accommodates both mean vector and covariance matrix for further performance improvement. We insert the Gaussian, identified as matrix square root of an SPD matrix, into deep CNNs based on Lie group theory \cite{L2EMG}. The resulting network, called G$^{2}$DeNet, is presented in Section~\ref{subsection:G2DeNet}. Secondly, forward propagation (FP) and backward propagation (BP) of both MPN-COV and G$^{2}$DeNet need to compute matrix power that depends on eigen-decomposition (EIG) or singular value decomposition (SVD). However, implementation of EIG/SVD is limitedly supported on CUDA platform, even much slower than their CPU counterparts \cite{Ionescu_2015_ICCV,lin2017improved}),  resulting in computational bottleneck. Inspired by~\cite{lin2017improved}, we develop an iterative matrix square root normalized covariance pooling (iSQRT-COV)\footnote{The matrix square root normalization is a special case of MPN-COV where the power is 1/2 that usually performs best among all values of power (Refer to experiments in Sec.~\ref{ablation}).} based on Newton-Schulz iteration \cite{Higham:2008:FM} for end-to-end learning, where we introduce key pre-normalization and post-compensation for the Newton-Schulz iteration, without which deeper CNNs (e.g., ResNet) fail to converge. The iSQRT-COV, as described in Section~\ref{iSQRT-COV},  is very suitable for parallel implementation on GPU, which can significantly speed up training of the networks. Thirdly, the output features of deep CNNs usually are of high dimension, resulting in much larger size of covariance representations that lead to high model complexity. To reduce model complexity without sacrificing performance, in Section~\ref{subsection: compact covariance}, we propose to exploit progressive $1\times1$ convolutions and group convolution \cite{XieGDTH17} to compress covariance representations. All these efforts bring about a unified methodology,  in which we not only improve significantly the existing GCP but also achieve fast training speed and affordable model complexity, making our final GCP networks very competitive and appealing.

The proposed GCP methods can be readily inserted into existing deep CNN architectures in an end-to-end manner. Fig.~\ref{overview} illustrates deep CNNs with our proposed MPN-COV and improved solutions. Finally, we conduct experiments on a variety of visual recognition tasks, including large-scale object classification, scene categorization, fine-grained visual recognition, and texture classification. This paper summarizes and extends our preliminary works~\cite{Wang_2016_CVPR,LiXWZ17,Wang_2017_CVPR,LiXWG18}, and the contributions are summarized as follows:

\begin{itemize}	[leftmargin=5mm]
\item[--] We propose a global Matrix Power Normalized COVariance (MPN-COV) pooling for deep CNNs, which can address the challenges of robust covariance estimation and usage of Riemannian geometry of covariances, further improving the representation and generalization abilities of deep CNNs.
\item[--] We propose several solutions to overcome the downsides of MPN-COV. First, we propose a Gaussian embedding network to properly incorporate additional first-order information. Then, we implement forward and backward propagations of MPN-COV with $\alpha=1/2$ based on Newton-Schulz iteration for fast training of MPN-COV networks. Additionally, a compact strategy of  progressive $1\times1$ convolutions and group convolution is introduced to reduce size of covariance representations. These solutions as well as MPN-COV constitute our matrix power normalization methodology for improving deep CNNs.
\item[--] The proposed methods are implemented on different deep learning platforms, and the complete code will be released to open source repository. We write C++ code based on NVIDIA \href{http://docs.nvidia.com/cuda/cublas/}{cuBLAS} and Matlab  using MatConvNet~\cite{vedaldi15matconvnet}. Meanwhile, we implement iSQRT-COV using PyTorch and TensorFlow packages. 
\item[--] Extensive evaluations are conducted on various visual recognition tasks, including large-scale object classification on ImageNet \cite{imagenet_cvpr09} and scene categorization on Places365 \cite{zhou2017places}, fine-grained visual recognition on Birds-CUB200-2011 \cite{WahCUB2002011}, Aircrafts \cite{aircraft} and Cars \cite{KrauseStarkDengFei-Fei_3DRR2013}, texture classification on DTD \cite{cimpoi14describing} and Indoor67 \cite{QuattoniT09}, and \href{https://www.kaggle.com/c/inaturalist-2018}{iNaturalist Challenge 2018} held in FGVC5 workshop in conjunction with CVPR 2018. The results with different CNN architectures (e.g., AlexNet \cite{nips2012cnn}, VGG-VD \cite{Simonyan15}, ResNet \cite{He_2016_CVPR} and DenseNet \cite{Huang_2017_CVPR}) show the superiority of our GCP networks.
\end{itemize}

\section{Related Work}
This section reviews the related works that improve deep CNNs by integration of trainable sophisticated pooling or encoding methods, which are divided into four categories. 

\subsection{Deep CNNs with Global Second-order Pooling}
Both DeepO$_{2}$P \cite{Ionescu_2015_ICCV} and B-CNN \cite{lin2015bilinear} insert a trainable second-order non-central moment into deep CNNs, where matrix logarithm normalization and element-wise power normalization followed by $\ell_{2}$-normalization are performed, respectively. Acharya et al.~\cite{Acharya_2018_CVPR_Workshops} explore a manifold network structure of covariance pooling for facial expression recognition, which performs, for input covariance matrix, bilinear mapping for dimension reduction, eigenvalue rectification and matrix logarithm  consecutively. In improved B-CNN~\cite{lin2017improved}, Lin and Maji study the effect of different normalization methods on second-order statistics, finding out matrix square root normalization offers significant improvement over other normalization methods. In particular, they for the first time propose to use Newton-Schulz iteration~\cite{Higham:2008:FM} for efficient FP while computing accurate gradients by Layapnov equation for BP.  While the idea of matrix square root normalization (a special case of MPN with power of 1/2) on GCP in improved B-CNN shares similarity with our MPN-COV~\cite{LiXWZ17},  we provide theoretical explanations on why  matrix power normalization works in deep architectures and experiment with large-scale ImageNet.  Our iSQRT-COV~\cite{LiXWG18}, which aims to speed up our MPN-COV, is inspired by improved B-CNN but has clear differences in several respects. First, we develop a meta-layer consisting  of pre-normalization, coupled matrix iteration and post-compensation, where pre-normalization (by trace or Frobenius norm) and post-compensation are distinctly important for convergence of deep CNNs (e.g., ResNet-50), and have been not studied in previous works including improved B-CNN. Second,  both FP and BP of iSQRT-COV are performed based on Newton-Schulz iteration, which is more efficient than improved B-CNN. Finally, our method outperforms improved B-CNN on both large-scale and small-scale classification, while making first attempt to show GCP can benefit deeper CNNs (e.g., ResNet-50).

\subsection{Deep CNNs with Global Approximate High-order Pooling}
In general, high-order pooling methods result in large-size representations. Gao et al. \cite{Gao_2016_CVPR} and Kong et al. \cite{Kong_Charless_2017_CVPR} propose compact B-CNN and low-rank B-CNN models to reduce sizes of covariance (second-order) representations, respectively. These methods replace exact covariances by small-size approximation ones, while achieving comparable performance. Based on compact B-CNN model \cite{Gao_2016_CVPR}, Dai et al. \cite{Dai_2017_CVPR} fuse additional first-order (mean) information by simply concatenating them. Kernel pooling \cite{Cui_2017_CVPR} extends the similar idea  with the compact B-CNN to approximate higher-order (number of order $> 2$) pooling. Sharing similar philosophy with \cite{Cui_2017_CVPR}, Cai et al. \cite{Cai_2017_ICCV} obtain small-size higher-order representations based on polynomial kernel approximation and rank-$1$ tensor decomposition \cite{KoldaB09}, namely HIHCA. Both kernel pooling \cite{Cui_2017_CVPR} and HIHCA \cite{Cai_2017_ICCV} improve compact B-CNN by exploiting higher-order  information. In~\cite{DBLP:conf/eccv/YuS18}, Yu and Salzmann propose a statistically-motivated second-order (SMSO) pooling, which successively consists of parametric vectorization, an element-wise square-root normalization, and a trainable affine transformation, each of which yields a well-defined distribution of data. SMSO can produce compact second-order representation while achieving state-of-the-art results. However, all above methods consider neither robust estimation nor geometry of manifold, limiting the ability of higher-order statistical modeling.

\subsection{Deep CNNs with Local Second-order Statistics}
In contrary to above works exploring global second-order pooling, some researchers try to incorporate local second-order statistics into deep CNNs. Among them,  Factorized Bilinear (FB) method \cite{LiYanghao_2017_ICCV} introduces an additional parametric quadratic term into linear transformation of convolution or fully-connected (FC) layers. FB can incorporate more complex non-linearity structures into deep CNNs by considering second-order interaction between information flow. Second-Order Response Transform (SORT) \cite{Wang_2017_ICCV} proposes to fuse outputs of two-branch block using a second-order term (i.e., element-wise product and a sum operation), in order to increase the nonlinearity of deep CNNs as well. Obviously, these methods also discard robust estimation and geometry of second-order statistics so that they cannot make full use of capacity of second-order pooling.

\subsection{Deep CNNs with Trainable BoVW Methods}
In the past decades, Bag-of-Visual-Words (BoVW) model is one of the most widely used orderless pooling methods for visual classification. Recently, some works insert high-performance BoVW methods \cite{JegouPDSPS12,sanchez} as trainable structural layers into deep CNNs. Thereinto, NetVLAD \cite{Arandjelovic_2016_CVPR} implements the modified vector of locally aggregated descriptors (VLAD) \cite{JegouPDSPS12} in an end-to-end manner. FisherNet \cite{TangWSBLT16} accomplishes the trainable layer of simplified Fisher vector (FV)~\cite{sanchez}. Unlike FisherNet, Li et al. \cite{LiYunSheng_2017_ICCV} propose a MFAFVNet for scene categorization, which performs a deep embedded implementation of mixture of factor analyzers Fisher vector (MFA-FV) method \cite{DixitV16}. Different from these methods, this paper proposes to integrate a global matrix power normalized covariance pooling into deep CNNs. 

\section{The Proposed Method}
In this section, we first describe our MPN-COV and the underlying mechanisms, and then instantiate our MPN-COV-Net. Subsequently, we present a global Gaussian embedding network to fuse first-order information, develop an iterative matrix square root normalization method to speed up training of networks, and introduce a compact strategy to reduce size of covariance representations.

\subsection{Global Matrix Power Normalized COVariance (MPN-COV) Pooling}\label{subsection:MPN-COV layer}
\subsubsection{Computation of MPN-COV}
Let $\mathcal{X}\in\mathbb{R}^{w\times h\times d}$ be the outputs of the convolution layer right before computation of MPN-COV, where $w$, $h$ and $d$ indicate spatial width, height and the number of channels, respectively. Then, the feature tensor $\mathcal{X}$ is reshaped to a feature matrix $\mathbf{X}\in\mathbb{R}^{d \times M}$ where $M=w \times h$. Given the feature matrix $\mathbf{X}$ consisting of $M$ samples with $d$-dimension, its sample covariance is computed as
\begin{align}\label{COV}
\boldsymbol{\Sigma} = \mathbf{X}\mathbf{J}\mathbf{X}^{T},\,\, \mathbf{J} = \frac{1}{M}(\mathbf{I}-\frac{1}{M}\mathbf{1}\mathbf{1}^{T}), 
\end{align}
where $\mathbf{I}$ indicates a $M \times M$ identity matrix, $\mathbf{1}$ is a $M$-dimension vector with all elements being one, and $T$ denotes the matrix transpose. 

Sample covariance $\boldsymbol{\Sigma}$ is a symmetric positive definite or semidefinite matrix, which can be factorized by EIG/SVD:
\begin{align}\label{equ:eig-cov}
\boldsymbol{\Sigma} = \mathbf{U}\boldsymbol{\Lambda}\mathbf{U}^{T},
\end{align}
where $\boldsymbol{\Lambda}=\mathrm{diag}(\lambda_{1},\ldots,\lambda_{d})$ is a diagonal matrix and $\lambda_{i}, i=1,\ldots, d$ are eigenvalues arranged in non-increasing order; $\mathbf{U}=[\mathbf{u}_{1},\ldots,\mathbf{u}_{d}]$ is an orthogonal matrix whose column $\mathbf{u}_{i}$ is the eigenvector corresponding to $\lambda_{i}$. Through EIG or SVD, we can compute matrix power as follows:
\begin{align}\label{equ:normlization-cov}
\mathbf{Z}\stackrel{\vartriangle}{=}\boldsymbol{\Sigma}^{\alpha}=\mathbf{U}\mathrm{diag}(f(\lambda_{1}),\ldots,f(\lambda_{d}))\mathbf{U}^{T}.
\end{align}
Here $\alpha > 0 $ is a scalar and $f(\lambda_{i})$ is  power of the eigenvalues
\begin{align}\label{equ:power-norm}
f(\lambda_{i})=\lambda_{i}^{\alpha}.
\end{align}

In this paper, the operation in Eq.~(\ref{equ:normlization-cov}) is called \emph{MPN-COV}, which is inserted after the last convolution layer of deep CNNs to collect second-order statistics of the output features (i.e., $\mathbf{X}$) as global representations. Next, we describe the mechanisms underlying MPN-COV in terms of robust covariance estimation and usage of geometry.

\begin{figure}
	\centering
	\includegraphics[width=1.0\linewidth]{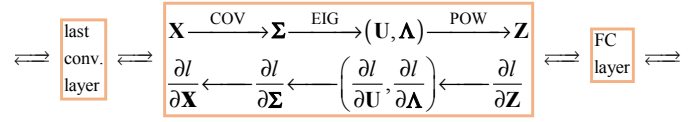}
	\caption{Diagram of the proposed MPN-COV block.}\label{fig:MPN-COV}
\end{figure}

\subsubsection{Robust Covariance Estimation}
Assuming $\mathbf{X}=[\mathbf{x}_1,\ldots,\mathbf{x}_{M}]$ are sampled from a Gaussian distribution, the covariance $\boldsymbol{\Sigma}$ of $\mathbf{X}$ can be estimated by optimizing the following objective function based on MLE: 
\begin{align}\label{MLE}
\arg\min_{\boldsymbol\Sigma} \log|\boldsymbol\Sigma| + \mathrm{tr}(\boldsymbol\Sigma^{-1}\mathbf{S}),
\end{align}
where $\mathbf{S} =\mathbf{X}\mathbf{J}\mathbf{X}^{T}$ is sample covariance, $|\cdot|$ indicates matrix determinant and $\mathrm{tr}(\cdot)$ means trace of matrix. The solution to the MLE~(\ref{MLE}) is the sample covariance, i.e., $\boldsymbol{\Sigma}=\mathbf{S}$. However, it is well known that MLE is not robust when data is of HDSS~\cite{Stein2001,Donoho2014}. This scenario is just what our covariance pooling faces: in most of the existing deep CNNs \cite{Simonyan15,Szegedy_2015_CVPR,He_2016_CVPR,Huang_2017_CVPR}, the output features (i.e., $\mathbf{X}$) of last convolution layer have less than 200 samples of dimension larger than 256, so the sample covariances are always ill-conditioned, rendering robust estimation critical.

For robust estimation of covariances in the case of HDSS, the general principle is shrinkage of eigenvalues of the sample covariances for counteracting the ill-conditioning of covariance matrices \cite{Stein1986,ledoit2004wellconditioned}. Besides, some researchers propose various regularized MLE methods for robust covariance estimation (see~\cite{icml2014c2_yangd14} and references therein). Notably, our MPN-COV closely conforms to the shrinkage principle~\cite{Stein1986,ledoit2004wellconditioned}, i.e., shrinking the largest sample eigenvalues and stretching the smallest ones. Moreover, MPN-COV can be deemed as a regularized MLE, namely vN-MLE ~\cite{Wang_2016_CVPR}:
\begin{align}\label{RMLE}
\arg\min_{\widehat{\boldsymbol\Sigma}} \log|\widehat{\boldsymbol\Sigma}| + \mathrm{tr}(\widehat{\boldsymbol\Sigma}^{-1}\mathbf{S}) + \gamma D_{\text{vN}}(\mathbf{I},\widehat{\boldsymbol\Sigma}),
\end{align}
where $\gamma>0$ is a regularizing constant, and  $D_{\text{vN}}(\mathbf{A},\mathbf{B})=\mathrm{tr}(\mathbf{A}(\log(\mathbf{A})-\log(\mathbf{B}))-\mathbf{A}+\mathbf{B})$ is matrix von-Neumann divergence. Compared with the classical MLE (\ref{MLE}) which only includes the first two terms, the objective function of vN-MLE~(\ref{RMLE}) introduces the third term, constraining the estimated covariance $\widehat{\boldsymbol\Sigma}$ be similar to the identity matrix $\mathbf{I}$.  In~\cite{Wang_2016_CVPR}, it has been shown that the vN-MLE outperforms other shrinkage methods~\cite{Stein1986,ledoit2004wellconditioned,ChenWEH10} and regularized MLE method~\cite{icml2014c2_yangd14}. Briefly, we have
\begin{proposition}\label{propostion-robust-estimation}
	MPN-COV with $\alpha=\frac{1}{2}$ is the unique solution to the vN-MLE in which  $\gamma=1$, i.e.,
	\begin{align}\label{equ:robust-estimation}
	\boldsymbol\Sigma^{\frac{1}{2}}=\arg\min_{\widehat{\boldsymbol\Sigma}} \log|{\widehat{\boldsymbol\Sigma}}| + \mathrm{tr}({\widehat{\boldsymbol\Sigma}}^{-1}{\mathbf{S}}) +  D_{\mathrm{vN}}(\mathbf{I},\widehat{\boldsymbol\Sigma}),
	\end{align}
	where $\boldsymbol\Sigma=\mathbf{S}$.
\end{proposition}
Proposition~\ref{propostion-robust-estimation} shows our MPN-COV with $\alpha=1/2$ performs robust estimation of covariance, and experiments in Section~\ref{ablation}  show that  $\alpha=1/2$ performs best. Details on vN-MLE can be referred to \cite{Wang_2016_CVPR}.

\subsubsection{Approximate Usage of Geometry}
Since the space of $d\times d$ covariance matrices (denoted by $\mathbb{S}^{+}_{d}$) forms a Riemannian manifold, geometry should be considered when distances between covariances are measured. There are mainly two kinds of Riemannian metrics, i.e., AIRM~\cite{Pennec2006} and LREM~\cite{Arsigny2005}. The AIRM is affine-invariant, but it is computationally inefficient and coupled, not scalable to large-scale scenarios. The LERM is a similarity-invariant decoupled metric and  efficient to compute, so that it is scalable to large-scale problems. Our MPN-COV can be regarded as matching covariance matrices with the Power-Euclidean (Pow-E) metric~\cite{Dryden2009}, which has close connection with the LERM, as presented in the following proposition:
\begin{proposition}\label{proposition-power-Euclidean}
	For any two covariance matrices $\boldsymbol\Sigma_{1}$ and $\boldsymbol\Sigma_{2}$, the limit of the Pow-E metric
	$d_{\alpha}(\boldsymbol\Sigma_{1},\boldsymbol\Sigma_{2})=\frac{1}{\alpha}\big\|\boldsymbol\Sigma^{\alpha}_{1}-\boldsymbol\Sigma_{2}^{\alpha}\big\|_{F}$
	as $\alpha>0$ approaches 0 equals the LERM, i.e.,
	$\lim\limits_{\alpha \rightarrow 0}d_{\alpha}(\boldsymbol\Sigma_{1},\boldsymbol\Sigma_{2})=\big\|\log(\boldsymbol\Sigma_{1})-\log(\boldsymbol\Sigma_{2})\big\|_{F}$.
\end{proposition}
This conclusion was first mentioned in~\cite{Dryden2009} but without proof. Here we briefly prove this proposition. Note that $d_{\alpha}(\boldsymbol\Sigma_{1},\boldsymbol\Sigma_{2})=\big\|\frac{1}{\alpha}(\boldsymbol\Sigma^{\alpha}_{1}-\mathbf{I})-\frac{1}{\alpha}(\boldsymbol\Sigma^{\alpha}_{2}-\mathbf{I})\big\|_{F}$. For any covariance $\boldsymbol\Sigma$ we have  $\frac{1}{\alpha}(\boldsymbol\Sigma^{\alpha}-\mathbf{I})=\mathbf{U}\mathrm{diag}(\frac{\lambda_{1}^{\alpha}-1}{\alpha},\ldots,\frac{\lambda_{n}^{\alpha}-1}{\alpha})\mathbf{U}^{T}$ based on its EIG. The identity about the limit in Proposition~\ref{proposition-power-Euclidean} follows immediately by recalling  $\lim_{\alpha\rightarrow 0}\frac{\lambda^{\alpha}-1}{\alpha}=\log(\lambda)$. Hence, the proposed MPN-COV can be viewed as approximately exploiting the Riemannian geometry of $\mathbb{S}^{+}_{d}$. It seems that the LERM is better than the Pow-E metric, since the former computes the true geodesic distance but the latter only measures it approximately. We argue that this is not the case for the scenario of deep CNNs from  the perspectives of both numerical stability and distribution of eigenvalues. Detailed discussion is given in Section~\ref{Comparisons}.

\subsection{Global MPN-COV Pooling  Network}\label{subsection: fp and bp}

We first instantiate a global matrix power normalized covariance pooling neural network (MPN-COV-Net) by inserting our MPN-COV after the last convolution layer of deep CNNs, in place of the common GAP. The diagram of our MPN-COV block is illustrated in Fig.~\ref{fig:MPN-COV}. For forward propagation, we first compute the covariance pooling of output features $\mathbf{X}$ of the last convolution layer using Eq.~(\ref{COV}), then we perform matrix power normalization (\ref{equ:normlization-cov}). Inspired by the element-wise power normalization technique~\cite{sanchez}, we can further perform, after MPN-COV, normalization by matrix $\ell_{2}-$norm (M-$\ell_{2}$) or by  matrix Frobenius norm (M-Fro). The matrix $\ell_{2}-$norm (also known as the \textit{spectral norm})  of a matrix  $\boldsymbol\Sigma$, denoted by $\|\boldsymbol\Sigma\|_{2}$, is defined as the largest singular value of $\boldsymbol\Sigma$, which equals the largest eigenvalue if $\boldsymbol\Sigma$ is a covariance matrix. The matrix Frobenius norm of $\boldsymbol\Sigma$ can be defined in various ways such as $\|\boldsymbol\Sigma\|_{F}=(\mathrm{tr}(\boldsymbol\Sigma^{T}\boldsymbol\Sigma))^{\frac{1}{2}}=(\sum_{i}\lambda_{i}^{2})^{\frac{1}{2}}$, where $\lambda_{i}$ are eigenvalues of $\boldsymbol\Sigma$. Then, we have
\begin{align}\label{equ:MPN-l2-or-fro}
\renewcommand*{\arraystretch}{1.2}
f(\lambda_{i})=\left\{ \begin{matrix}
{\lambda_{i}^{\alpha}}\Big/{\lambda_{1}^{\alpha}} & \text{for MPN-COV+M-$\ell_{2}$} \\
{\lambda_{i}^{\alpha}}\Big/(\sum_{k} \lambda_{k}^{2\alpha})^{\frac{1}{2}} & \text{for MPN-COV+M-Fro}
\end{matrix}\right.
\end{align}

For backpropagation of MPN-COV block, we need to compute the partial derivative of loss function $l$ with respect to the input $\mathbf{X}$ based on the methodology of matrix backpropagation~\cite{Ionescu_2015_ICCV,IonescuVS15}. First of all, given $\frac{\partial l}{\partial \mathbf{Z}}$ propagated from  top  layer, we compute the derivatives  $\frac{\partial l}{\partial \mathbf{U}}$ and $\frac{\partial l}{\partial \boldsymbol{\Lambda}}$ based on the following chain rule:
\begin{align}\label{equ:chain-rule}
\mathrm{tr}\Big(\Big(\frac{\partial l}{\partial \mathbf{U}}\Big)^{T}\mathrm{d}\mathbf{U}+\Big(\frac{\partial l}{\partial \boldsymbol{\Lambda}}\Big)^{T}\mathrm{d}\boldsymbol{\Lambda}\Big)=\mathrm{tr}\Big(\Big(\frac{\partial l}{\partial \mathbf{Z}}\Big)^{T}\mathrm{d}\mathbf{Z}\Big),
\end{align}
where $\mathrm{d}\mathbf{Z}$ denotes variation of matrix $\mathbf{Z}$. According to Eq.~ (\ref{equ:normlization-cov}), we have $\mathrm{d}\mathbf{Z}=\mathrm{d}\mathbf{U}\mathbf{F}\mathbf{U}^{T}+\mathbf{U}\mathrm{d}\mathbf{F}\mathbf{U}^{T}+\mathbf{U}\mathbf{F}\mathrm{d}\mathbf{U}^{T}$, where  $\mathbf{F}=\mathrm{diag}(f(\lambda_{1}),\ldots,f(\lambda_{d}))$ and  $\mathrm{d}\mathbf{F}=\mathrm{diag}\big(\alpha \lambda_{1}^{\alpha-1}, \ldots, \alpha \lambda_{d}^{\alpha-1}\big)\mathrm{d}\boldsymbol{\Lambda}$. After some arrangements, we obtain
\begin{align}\label{equ:backward_step-norm}
\dfrac{\partial l}{\partial \mathbf{U}}&= \Big(\dfrac{\partial l}{\partial \mathbf{Z}}+\Big(\dfrac{\partial l}{\partial \mathbf{Z}}\Big)^{T}\Big)\mathbf{U}\mathbf{F},\\
\dfrac{\partial l}{\partial \boldsymbol{\Lambda}}&=\alpha\Big(\mathrm{diag}\Big( \lambda_{1}^{\alpha-1}, \ldots,  \lambda_{d}^{\alpha-1}\Big)\mathbf{U}^{T}\dfrac{\partial l}{\partial \mathbf{Z}}\mathbf{U}\Big)_{\mathrm{diag}},\nonumber
\end{align}
where $(\cdot)_{\mathrm{diag}}$ denotes the matrix diagonalization. For MPN-COV+M-$\ell_{2}$ and MPN-COV+M-Fro, $\frac{\partial l}{\partial \boldsymbol{\Lambda}}$ takes respectively the following forms:
\begin{align}\label{equ:backward_step-power_l2}
\dfrac{\partial l}{\partial \boldsymbol{\Lambda}}=&\dfrac{\alpha}{\lambda_{1}^{\alpha}}\Big(\mathrm{diag}\Big({\lambda_{1}^{\alpha-1}}, \ldots, {\lambda_{d}^{\alpha-1}}\Big)\mathbf{U}^{T}\dfrac{\partial l}{\partial \mathbf{Z}}\mathbf{U}\Big)_{\mathrm{diag}}\\
&- \mathrm{diag}\bigg(\dfrac{\alpha}{\lambda_{1}}\mathrm{tr}\Big(\mathbf{Z}\dfrac{\partial l}{\partial \mathbf{Z}}\Big), 0, \ldots, 0\bigg)\nonumber
\end{align}
and
\begin{align}\label{equ:backward_step-power_fro}
\dfrac{\partial l}{\partial \boldsymbol{\Lambda}}=&\dfrac{\alpha}{\sqrt{\sum\nolimits_{k} \lambda_{k}^{2\alpha}}}\Big(\mathrm{diag}\Big({ \lambda_{1}^{\alpha-1}}, \ldots, { \lambda_{d}^{\alpha-1}}\Big)\mathbf{U}^{T}\dfrac{\partial l}{\partial \mathbf{Z}}\mathbf{U}\Big)_{\mathrm{diag}}\nonumber\\
-&\dfrac{\alpha}{{\sum\nolimits_{k} \lambda_{k}^{2\alpha}}}\mathrm{tr}\Big(\mathbf{Z}\dfrac{\partial l}{\partial \mathbf{Z}}\Big)\mathrm{diag}\Big({ \lambda_{1}^{2\alpha-1}}, \ldots, {\lambda_{d}^{2\alpha-1}}\Big).
\end{align}

Next, given  $\frac{\partial l}{\partial \mathbf{U}}$ and $\frac{\partial l}{\partial \boldsymbol{\Lambda}}$, we need to compute $\frac{\partial l}{\partial \boldsymbol\Sigma}$  associated with Eq.~(\ref{equ:eig-cov}), whose corresponding  chain rule is $\mathrm{tr}((\frac{\partial l}{\partial \boldsymbol\Sigma})^{T}\mathrm{d}\boldsymbol\Sigma)=\mathrm{tr}((\frac{\partial l}{\partial \mathbf{U}})^{T}\mathrm{d}\mathbf{U}+(\frac{\partial l}{\partial\boldsymbol{\Lambda} })^{T}\mathrm{d}\boldsymbol{\Lambda})$. Note that $\mathbf{U}$ is an orthogonal matrix. After some arrangements, we have
\begin{align}\label{equ:eigendecomposiiton-backward}
\dfrac{\partial l}{\partial \boldsymbol\Sigma}=\mathbf{U}\Big(\Big(\mathbf{K}^{T}\circ \Big(\mathbf{U}^{T}\dfrac{\partial l}{\partial \mathbf{U}}\Big)\Big)+\Big(\dfrac{\partial l}{\boldsymbol{\partial \Lambda}}\Big)_{\mathrm{diag}}\Big)\mathbf{U}^{T},
\end{align}
where $\circ$ denotes matrix Hadamard product. The matrix $\mathbf{K}=\{K_{ij}\}$ where $K_{ij}=1/(\lambda_{i}-\lambda_{j})$ if $i\neq j$ and $K_{ij}=0$ otherwise. We refer readers to ~\cite[Proposition 2]{IonescuVS15} for in-depth derivation of Eq.~(\ref{equ:eigendecomposiiton-backward}).

Finally, given $\frac{\partial l}{\partial \boldsymbol\Sigma}$, we derive the gradient of the loss function with respect to the input feature $\mathbf{X}$, which has
\begin{align}\label{equ:compute-cov-backward}
\dfrac{\partial l}{\partial \mathbf{X}}=\bigg(\dfrac{\partial l}{\partial \boldsymbol\Sigma}+\bigg(\dfrac{\partial l}{\partial \boldsymbol\Sigma}\bigg)^{T}\bigg)\mathbf{X}\mathbf{J}.
\end{align}
As described above, by using the formulas of Eq.~(\ref{equ:normlization-cov}) and Eq.~(\ref{equ:compute-cov-backward}), our MPN-COV-Net can be end-to-end trained.
 
\subsection{Global Gaussian Embedding Network}\label{subsection:G2DeNet}

Our MPN-COV can address the two challenges GCP  faces, clearly outperforming other competing methods. Nevertheless, it neglects the first-order representation (i.e., mean vector) that is widely used in conventional deep CNNs. Previous researches~\cite{sanchez,WangLZZ16,Wang_2016_CVPR} have shown that combination of first- and second-order statistics is often better than either one single statistics. As is commonly done, we use Gaussians for modeling distributions of features.  However, as the space of Gaussians is a  manifold, it is challenging to insert a Gaussian into deep CNNs. As such, we propose to use Gaussian embedding method proposed in~\cite{L2EMG}.  This embedding method  equips the space of Gaussians with a Lie group structure, respecting both geometrical  and algebraic structures, and meanwhile it identifies a Gaussian as matrix square root of an SPD matrix, suitable for backpropagation while achieving better performance. In Appendix II, we compare with other embedding methods~\cite{Calvo,RePEcjmvana,GongWang2009,Nakayama-CVPR2010,L2EMG}.   In the following, we first describe Gaussian embedding method and then present FP and BP of the resulting network (i.e., G$^2$DeNet).

\textbf{Gaussian Embedding} Let $\mathbf{X}=[\mathbf{x}_{1},\ldots,\mathbf{x}_{M}] \in \mathbb{R}^{d\times M}$ be a set of deep convolution features, we use Gaussian to model feature distribution as 
\begin{align*}
	p(\mathbf{x}) = \frac{1}{(2\pi)^{\frac{d}{2}}|\boldsymbol\Sigma|^{\frac{1}{2}}} \exp \big(-\frac{1}{2}(\mathbf{x}-\boldsymbol\mu)^{T}\boldsymbol\Sigma^{-1}(\mathbf{x}-\boldsymbol\mu)\big),
	\end{align*}
	where $\boldsymbol{\mu}=\frac{1}{M} \sum_{i=1}^{M} \mathbf{x}_{i}$ and  $\boldsymbol{\Sigma}=\frac{1}{M}\sum_{i=1}^{M}(\mathbf{x}_{i}-\boldsymbol\mu)(\mathbf{x}_{i}-\boldsymbol\mu)^{T}$ are  mean vector and sample covariance matrix, respectively.  Let  $\boldsymbol{\Sigma}^{-1}=\mathbf{L}\mathbf{L}^{T}$ be the Cholesky decomposition of the inverse of $\boldsymbol{\Sigma}$, where $\mathbf{L}$ is a lower triangular matrix of order $d$ with positive diagonals. The Gaussian $\mathcal{N}(\boldsymbol{\mu},\boldsymbol{\Sigma})$ can be uniquely mapped to a positive definite upper triangular matrices of order $d+1$ through the mapping $\phi$:
\begin{align}\label{map1}
\phi:\mathcal{N}(\boldsymbol{\mu},\boldsymbol{\Sigma})\xrightarrow{\boldsymbol{\Sigma}^{-1}=\mathbf{L}\mathbf{L}^{T}}\mathbf{H}_{\boldsymbol{\mu},\mathbf{J}}\stackrel{\vartriangle}{=}\left[\begin{matrix} \mathbf{J} & \boldsymbol{\mu} \\ \mathbf{0}^{T} & 1 \end{matrix}\right],
\end{align}
where $\mathbf{J}=\mathbf{L}^{-T}$ and $\mathbf{H}_{\boldsymbol{\mu},\mathbf{J}}\in UT^{+}(d+1)$. $UT^{+}(d+1)$ indicates the set of all positive definite upper triangular matrices of order $d+1$. We note that the embedding  (\ref{map1}) is not suitable for backpropagation due to Cholesky decomposition and matrix inverse. 
The matrix $\mathbf{H}_{\boldsymbol{\mu},\mathbf{J}}$ can be further mapped to a unique SPD matrix through a mapping $\psi$ based on its matrix polar decomposition  $\mathbf{H}_{\boldsymbol{\mu},\mathbf{J}}=\mathbf{S}_{\boldsymbol{\mu},\mathbf{J}}\mathbf{Q}_{\boldsymbol{\mu},\mathbf{J}}$, i.e.,  
\begin{align}\label{equ:SQ}
\psi: \mathbf{H}_{\boldsymbol{\mu},\mathbf{J}} \xrightarrow{\mathbf{H}_{\boldsymbol{\mu},\mathbf{J}}=\mathbf{S}_{\boldsymbol{\mu},\mathbf{J}}\mathbf{Q}_{\boldsymbol{\mu},\mathbf{J}}} \mathbf{S}_{\boldsymbol{\mu},\mathbf{J}},
\end{align}
where $\mathbf{S}_{\boldsymbol{\mu},\mathbf{J}}$ is an SPD matrix and $\mathbf{Q}_{\boldsymbol{\mu},\mathbf{J}}$ is the closest orthogonal matrix to $\mathbf{H}_{\boldsymbol{\mu},\mathbf{J}}$. 
Through the two consecutive mappings $\phi$ and $\psi$, we  identify a Gaussian as matrix square root of an SPD matrix:
\begin{align}\label{RGE}
\mathcal{N}(\boldsymbol{\mu},\boldsymbol{\Sigma}) \xrightarrow{\psi \circ \phi} \mathbf{S}_{\boldsymbol{\mu},\mathbf{J}}=\begin{bmatrix}
\boldsymbol{\Sigma}+\boldsymbol{\mu}\boldsymbol{\mu}^{T} & \boldsymbol{\mu}\\
\boldsymbol{\mu}^{T} & 1
\end{bmatrix}^{\frac{1}{2}}.
\end{align}
We suggest readers  refer to \cite{L2EMG} for theoretical details.

\begin{figure}
	\centering
	\includegraphics[width=0.9\linewidth]{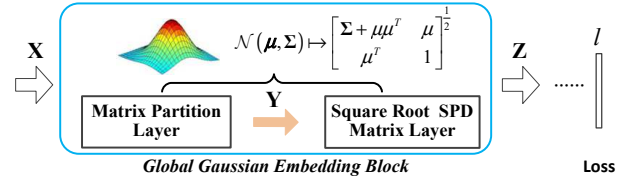}
	\caption{Diagram of the proposed global Gaussian embedding block.}\label{fig:G2DeNet}
\end{figure}

\textbf{FP and BP of G$^2$DeNet} According to the embedding form in Eq.~(\ref{RGE}), we design global Gaussian embedding block which consists of two layers, i.e., \emph{matrix partition layer} and \emph{square root SPD matrix layer}, as shown in  Fig.~\ref{fig:G2DeNet}. 
Below we first describe the \emph{matrix partition layer}.  We let  $\mathbf{Y}=\mathbf{S}_{\boldsymbol{\mu},\mathbf{J}}^{2}$, in which the mean vector $\boldsymbol{\mu}$ and covariance matrix $\boldsymbol{\Sigma}$ are obviously entangled. The purpose of this layer is to decouple $\mathbf{Y}$ and explicitly write it as the function of input feature matrix $\mathbf{X}$ to facilitate  computation of the derivatives. We  note that there exists the identity  $\boldsymbol{\Sigma}=\frac{1}{M}\mathbf{X}\mathbf{X}^{T}-\boldsymbol{\mu}\boldsymbol{\mu}^{T}$. After some elementary manipulations, we have
\begin{align}\label{MPL}
\mathbf{Y} =  \frac{1}{M}\mathbf{A}\mathbf{X}\mathbf{X}^{T}\mathbf{A}^{T}+ \frac{2}{M}\left(\mathbf{A}\mathbf{X}\mathbf{1}\mathbf{b}^{T}\right)_{\mathrm{sym}}+\begin{bmatrix}
\mathbf{O} & \mathbf{0} \\ \mathbf{0}^{T} & 1 \end{bmatrix}
\end{align}
where $\mathbf{A}=\begin{bmatrix} \mathbf{I} & \mathbf{0} \end{bmatrix}^{T}$ consists of a $d \times d$  identity matrix $\mathbf{I}$ and a $d-$dimensional zero vector $\mathbf{0}$,  $\mathbf{b}$ is a $(d+1)-$dimensional vector with all elements being zero except the last one which is equal to one,  $\mathbf{1}$ is a $M-$dimensional vector with all elements being one, and $\mathbf{O}$ is a $d\times d$ zero matrix. The notation $(\mathbf{F})_{\mathrm{sym}}=\frac{1}{2}\left(\mathbf{F}+\mathbf{F}^{T}\right)$ denotes matrix symmetrization.

The purpose of \emph{Square Root SPD Matrix Layer} is to compute the matrix square root of SPD matrix $\mathbf{Y}$, i.e.,  $\mathbf{Z} = \mathbf{Y}^{\frac{1}{2}}$. Similar to matrix power as described previously, the matrix $\mathbf{Z}$ can be computed via EIG/SVD.  It is straightforward to know that  $\mathbf{Y}$
is an SPD matrix, and so it can be factorized as  
\begin{equation}\label{equ:eigen-decomposition}
\mathbf{Y}=\widetilde{\mathbf{U}}\widetilde{\boldsymbol{\Lambda}}\widetilde{\mathbf{U}}^{T},
\end{equation} where $\widetilde{\boldsymbol{\Lambda}}=\mathrm{diag}(\widetilde{\lambda}_{1},\cdots,\widetilde{\lambda}_{d+1})$ and $\widetilde{\mathbf{U}}=[\widetilde{\mathbf{u}}_{1}\; \cdots\; \widetilde{\mathbf{u}}_{d+1}]$ are the eigenvalues and eigenvectors of $\mathbf{Y}$, respectively. The matrix square root of $\mathbf{Y}$ can be computed as:
\begin{align}\label{ESRL}
\mathbf{Z} =\widetilde{\mathbf{U}}\widetilde{\boldsymbol{\Lambda}}^{\frac{1}{2}}\widetilde{\mathbf{U}}^{T},
\end{align}
where $\widetilde{\boldsymbol{\Lambda}}^{\frac{1}{2}}=\mathrm{diag}(\widetilde{\lambda}_{1}^{\frac{1}{2}},\cdots,\widetilde{\lambda}_{d+1}^{\frac{1}{2}})$ is computed as element-wise square root of the eigenvalues. Combining matrix partition layer (\ref{MPL}) with square root SPD matrix layer (\ref{ESRL}), we can accomplish FP of  global Gaussian embedding block.

Given $\frac{\partial l}{\partial \mathbf{Z}}$ propagated from  top  layer,  we perform  BP   by computing $\frac{\partial l}{\partial\mathbf{Y}}$ and $\frac{\partial l}{\partial\mathbf{X}}$ associated with square root SPD matrix layer (\ref{ESRL}) and matrix partition layer (\ref{MPL}), respectively. Note that derivation of $\frac{\partial l}{\partial\mathbf{Y}}$ shares similar philosophy with that of Eq.~(\ref{equ:eigendecomposiiton-backward}), and we omit it  for simplicity. Given $\frac{\partial l}{\partial\mathbf{Y}}$, we can compute $\frac{\partial l}{\partial\mathbf{X}}$ based on Eq.~(\ref{MPL}), which takes the following form:
\begin{align}\label{final}
&\frac{\partial l}{\partial\mathbf{X}} = \frac{2}{M}\mathbf{A}^{T}\Big(\frac{\partial l}{\partial\mathbf{Y}}\Big)_{\mathrm{sym}}
\Big(\mathbf{A}\mathbf{X}+\mathbf{b}\mathbf{1}^{T}\Big).
\end{align}

\subsection{Iterative Matrix Square Root Normalized Covariance Pooling Network}\label{iSQRT-COV}
In our MPN-COV-Net and G$^2$DeNet, computation of matrix square root heavily depends on EIG or SVD. However, fast implementation of EIG or SVD on GPU is still an open problem. Table~\ref{table:time-matrix-decomposition} presents the running time of EIG and SVD of a $256\times 256$ covariance matrix. Matlab~(M) built-in CPU functions deliver over 10x and 5x speedups over the CUDA counterparts and built-in GPU functions, respectively. As such, both our MPN-COV-Net and G$^2$DeNet opt for EIG or SVD on CPU for computing matrix square root, greatly restricting the training speed of the networks on GPUs. 

\begin{table}[thb]
	\centering
	\footnotesize
	\caption{Time (ms) taken by matrix decomposition (single precision arithmetic) of a $256\times 256$ covariance matrix.}
	\label{table:time-matrix-decomposition}
	\begin{tabular}{c|c|c|c}
		\hline
		Algorithm                            & \parbox{0.50in}{\centering CUDA\\cuSOLVER}    & \parbox{0.8in}{\centering \vspace{2pt} Matlab\\(CPU function)\vspace{2pt}} & \parbox{0.8in}{\centering Matlab\\(GPU function)} \\
		\hline
		EIG                                  &    21.3 & 1.8    & 9.8     \\
		SVD                                  &    52.2 & 4.1    & 11.9     \\
		\hline
	\end{tabular}
\end{table}
	
\begin{figure}
	\centering
	\includegraphics[width=1.0\linewidth]{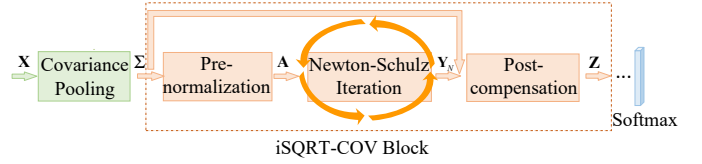}
	\caption{Diagram of the proposed iSQRT-COV block.}\label{fig:iSQRT-COV}
\end{figure}

To overcome this limitation, inspired by~\cite{lin2017improved}, we develop a fast training method, which is called iterative matrix square root normalization of covariance (iSQRT-COV) pooling, making use of Newton-Schulz iteration \cite{Higham:2008:FM} in both FP and BP. At the core of iSQRT-COV is a meta-layer with loop-embedded directed graph  structure, which consists of three consecutive  structured layers, performing pre-normalization, coupled matrix iteration and post-compensation, respectively. The iSQRT-COV block is illustrated in Fig.~\ref{fig:iSQRT-COV}, and the details are described as follows.

\textbf{Newton-Schulz Iteration}
Higham~\cite{Higham:2008:FM} studied a class of methods for iteratively computing  matrix square root. These methods, termed as  Newton-Pad\'{e} iterations, are developed based on the connection between  matrix sign function and matrix square root, together with  rational Pad\'{e} approximation.  Specifically, for computing the matrix square root of $\mathbf{A}$, given $\mathbf{Y}_{0}=\mathbf{A}$ and $\mathbf{P}_{0}=\mathbf{I}$, for $k=1,\cdots, N$, the coupled iteration takes the following form~\cite[Chap. 6.7]{Higham:2008:FM}:
\begin{align}\label{equ:coupled-equation0}
\mathbf{Y}_{k}&=\mathbf{Y}_{k-1}p_{lm}(\mathbf{P}_{k-1}\mathbf{Y}_{k-1})q_{lm}(\mathbf{P}_{k-1}\mathbf{Y}_{k-1})^{-1}\nonumber \\
\mathbf{P}_{k}&=p_{lm}(\mathbf{P}_{k-1}\mathbf{Y}_{k-1})q_{lm}(\mathbf{P}_{k-1}\mathbf{Y}_{k-1})^{-1}\mathbf{P}_{k-1},
\end{align}
where $p_{lm}$ and $q_{lm}$ are polynomials, and $l$ and $m$ are non-negative integers. According to Eq.~(\ref{equ:coupled-equation0}), $\mathbf{Y}_{k}$ and $\mathbf{P}_{k}$ quadratically converge to $\mathbf{A}^{1/2}$ and $\mathbf{A}^{-1/2}$, respectively. However, it converges only locally, i.e., it converges if $\|\mathbf{I}-\mathbf{A}\|<1$ where  $\|\cdot\|$  denotes any induced (or consistent) matrix norm. The family of coupled iteration is  stable in that small errors in the previous iteration will not be amplified.  The case of $l=0, m=1$ called \textit{Newton-Schulz iteration}   fits for our purpose as no GPU unfriendly matrix inverse is involved:
\begin{align}\label{equ:coupled-equation}
\mathbf{Y}_{k}&=\dfrac{1}{2}\mathbf{Y}_{k-1}(3\mathbf{I}-\mathbf{P}_{k-1}\mathbf{Y}_{k-1})\nonumber \\
\mathbf{P}_{k}&=\dfrac{1}{2}(3\mathbf{I}-\mathbf{P}_{k-1}\mathbf{Y}_{k-1})\mathbf{P}_{k-1}, 
\end{align}
where $k=1,\ldots,N$. Clearly Eq. (\ref{equ:coupled-equation}) involves only matrix product, suitable for parallel implementation on GPU. Compared to \textit{accurate} square root computed by EIG, one can obtain \textit{approximate} solution with a small number of iterations $N$, which is determined by cross-validation.

\textbf{Pre-normalization and Post-compensation} As Newton-Schulz iteration only converges locally, we pre-normalize covariance $\boldsymbol{\Sigma}$ by its trace or Frobenius norm, i.e.,
\begin{align}\label{equ:pre-normalization}
\mathbf{A}=\frac{1}{\mathrm{tr}(\boldsymbol{\Sigma})}\boldsymbol{\Sigma} \;\; \text{or}\;\;
\frac{1}{\|\boldsymbol{\Sigma}\|_{F}}\boldsymbol{\Sigma}.
\end{align}
Let $\lambda_{i}$ be eigenvalues of $\boldsymbol{\Sigma}$, arranged in nondecreasing order. As $\mathrm{tr}(\boldsymbol{\Sigma})=\sum_{i}\lambda_{i}$ and $\|\boldsymbol{\Sigma}\|_{F}=\sqrt{\sum_{i}\lambda_{i}^{2}}$, it is easy to see that $\|\mathbf{I}-\mathbf{A}\|_{2}$, which equals to the largest singular value of $\mathbf{I}-\mathbf{A}$, is $1-\frac{\lambda_{1}}{\sum_{i}\lambda_{i}}$ and $1-\frac{\lambda_{1}}{\sqrt{\sum_{i}\lambda_{i}^{2}}}$ for the case of trace and Frobenius norm, respectively, both less than 1. Hence, the convergence condition is satisfied. 

The above pre-normalization for covariance pooling nontrivially changes the data magnitudes, which produces side effect on network such that the prevalent ResNet~\cite{He_2016_CVPR} fails to converge. To counteract this adverse influence,  after the Newton-Schulz iteration,  we accordingly perform post-compensation as follows: 
\begin{align}\label{equ:post-compensation}
\mathbf{Z}=\sqrt{\mathrm{tr}(\boldsymbol{\Sigma)}}\mathbf{Y}_{N}\;\; \text{or}\;\; \mathbf{Z}=\sqrt{\|\boldsymbol{\Sigma}\|_{F}}\mathbf{Y}_{N}.
\end{align}
Alternatively, one may consider Batch Normalization (BN)~\cite{icml2015_ioffe15}, which does work, successfully helping ResNet convergence. Compared to BN, our post-compensation achieves about 1\% lower top-1 error with ResNet-50 architecture on ImageNet (see Section~\ref{ablation} for details).

Note that derivation of BP of iSQRT-COV is not straightforward. Despite autograd toolkits provided by some deep learning frameworks can accomplish this task automatically, the involved BP is still in a black box. Meanwhile, autograd sometimes brings uncertainty, e.g., autograd of PyTorch 0.3.0 or below cannot compute gradients of iSQRT-COV correctly. To make iSQRT-COV be self-contained and enable its implementation to be accessible when autograd toolkit is unavailable (e.g., the well-known~\href{http://caffe.berkeleyvision.org/}{Caffe} and early versions of~\href{http://www.vlfeat.org/matconvnet/}{MatConvNet}), we show the gradient of the loss function $l$ with respect to $\boldsymbol{\Sigma}$ and give the derivation details in the Appendix I for conciseness. Specifically, given $\frac{\partial l}{\partial \mathbf{Z}}$, we can compute the gradient for pre-normalization by trace as 
\begin{align}\label{equ:BP-pre-trace}
\dfrac{\partial l}{\partial \boldsymbol{\Sigma}}=&-\dfrac{1}{({\mathrm{tr}(\boldsymbol{\Sigma)}})^2}\mathrm{tr}\Big(\Big(\dfrac{\partial l}{\partial \mathbf{A}}\Big)^{T}\boldsymbol{\Sigma}\Big)\mathbf{I}+\dfrac{1}{\mathrm{tr}(\boldsymbol{\boldsymbol{\Sigma}})}\dfrac{\partial l}{\partial \mathbf{A}}\nonumber\\
&+\dfrac{1}{2\sqrt{\mathrm{tr}(\boldsymbol{\Sigma)}}}\mathrm{tr}\Big(\Big(\dfrac{\partial l}{\partial \mathbf{Z}}\Big)^{T}\mathbf{Y}_{N}\Big)\mathbf{I}.
\end{align}
If pre-normalization by Frobenius norm is adopted, we have
\begin{align}\label{equ:BP-pre-fro}
\dfrac{\partial l}{\partial \boldsymbol{\Sigma}}=&-\dfrac{1}{\|\boldsymbol{\Sigma}\|_{F}^{3}}\mathrm{tr}\Big(\Big(\dfrac{\partial l}{\partial \mathbf{A}}\Big)^{T}\boldsymbol{\Sigma}\Big)\boldsymbol{\Sigma}+\dfrac{1}{\|\boldsymbol{\Sigma}\|_{F}}\dfrac{\partial l}{\partial \mathbf{A}}\nonumber\\
&+\dfrac{1}{2\|\boldsymbol{\Sigma}\|_{F}^{3/2}}\mathrm{tr}\Big(\Big(\dfrac{\partial l}{\partial \mathbf{Z}}\Big)^{T}\mathbf{Y}_{N}\Big)\boldsymbol{\Sigma},
\end{align}
where $\frac{\partial l}{\partial \mathbf{A}}$ can be found in Appendix I. Finally, given $\frac{\partial l}{\partial \boldsymbol{\Sigma}}$,  the gradient of  $l$ with respect to  $\mathbf{X}$ is given in Eq.~(\ref{equ:compute-cov-backward}).

\subsection{Compact Covariance Representations}\label{subsection: compact covariance}
\begin{figure}
	\centering
	\includegraphics[width=1.0\linewidth]{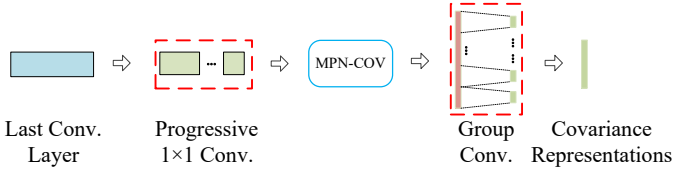}
	\caption{Diagram of the proposed compact strategy. We propose to combine \textbf{progressive $1\times1$ convolutions} and \textbf{group convolution} to compress covariance representations, as indicated by red dotted boxes.}\label{fig:groupconv}
\end{figure}

Given an input feature matrix $\mathbf{X} \in \mathbb{R}^{d\times M}$ consisting of $M$ samples with $d$-dimension, size of covariance representation is $d\times(d+1)/2$ after matrix vectorization by considering the symmetry. Taking $d=512$ or $d=1024$ as an example, the size of covariance representation is $131\text{,}328$ or $524\text{,}800$. Such large-size representation leads to high model complexity. To deal with this problem, we present a compact strategy for compressing covariance representations, as shown in Fig.~\ref{fig:groupconv}. The compact strategy consists of progressive $1\times1$ convolutions and group convolution. The formers are used to reduce feature dimension right before computation of covariance matrices, while the latter is employed to reduce size of covariance representations after GCP block.

Since the dimension of input features decides size of covariance representation, we first introduce a strategy of progressive $1\times1$ convolutions to reduce the dimension of input features. The $1\times1$ convolution is first introduced in \cite{iclr2014_NIN}, and has been widely used in deep CNN architectures \cite{Szegedy_2015_CVPR,He_2016_CVPR,Huang_2017_CVPR} for parameter reduction. However, an abrupt dimensionality reduction (e.g., directly from $2080$ to $128$) may hurt the performance of covariance representations. Therefore, we propose to progressively reduce the dimension of input features from $d$ to $\hat{d}$ ($\hat{d} \ll d$) using a set of consecutive $1\times1$ convolutions, i.e., $d=d_0 \rightarrow d_1 \rightarrow \dots \rightarrow d_K=\hat{d}$ with $d_{k-1}>d_{k}$. In this way, dimension of covariance representation can be effectively reduced to $\hat{d}\times(\hat{d}+1)/2$ (e.g., $8256$ for $\hat{d}=128$). Compared with dimensionality reduction using only one $1\times1$ convolution, our progressive convolutions can achieve better performance. 

Once dimension of input is determined, we can further reduce the size of covariance representation by learning a low-dimensional linear transform matrix $\mathbf{W}$ on covariance representation $\mathbf{z}$, which corresponds to inserting a FC layer ($\mathbf{\hat{z}} = \mathbf{W}\mathbf{z}+\mathbf{b}$) after covariance pooling. The number of parameters of $\mathbf{W}$ is $d_{\mathbf{z}}\times d_{\mathbf{\hat{z}}}$, where  $d_{\mathbf{z}}$ and $d_{\mathbf{\hat{z}}}$ are dimensions of $\mathbf{z}$ and $\mathbf{\hat{z}}$, respectively. If $d_{\mathbf{z}}$ is very high, transform matrix $\mathbf{W}$ will involve of many parameters, increasing computation and memory costs. Therefore, we introduce group convolution \cite{XieGDTH17} to alleviate this problem. Different from standard convolution, group convolution divides $\mathbf{z}$ into $G$ groups and performs convolution operation in each sub-group. As such, number of parameters of $\mathbf{W}$ can be decreased to  $(d_{\mathbf{z}}\times d_{\mathbf{\hat{z}}})/G$, very suitable for large-size covariances.

\begin{figure*}
	\centering
	\subfigure[]{
		\centering
		\includegraphics[width=0.31\textwidth]{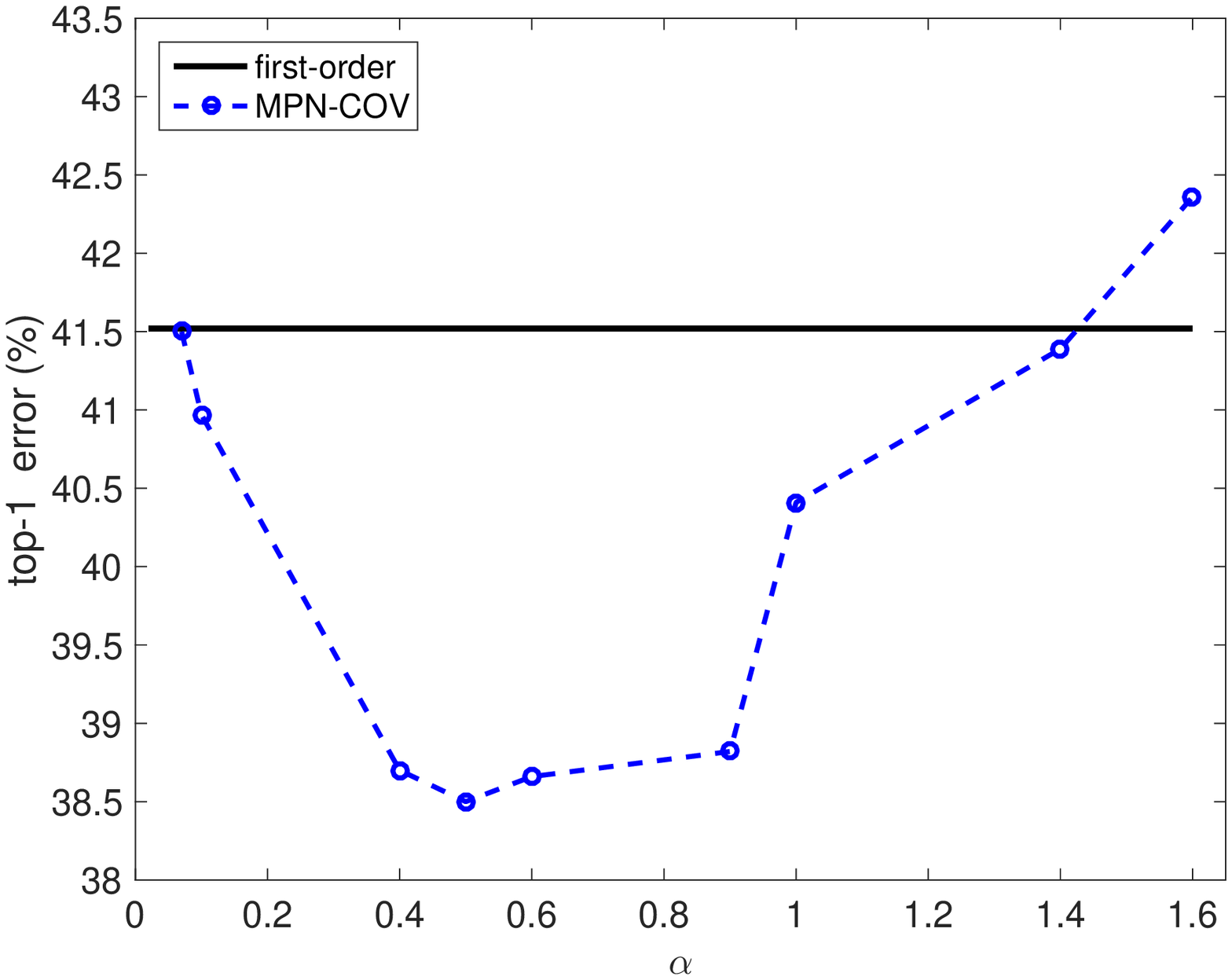} 
		\label{fig:impact_alpha} 
	}
	\subfigure[]{
		\centering
		\includegraphics[width=0.31\textwidth]{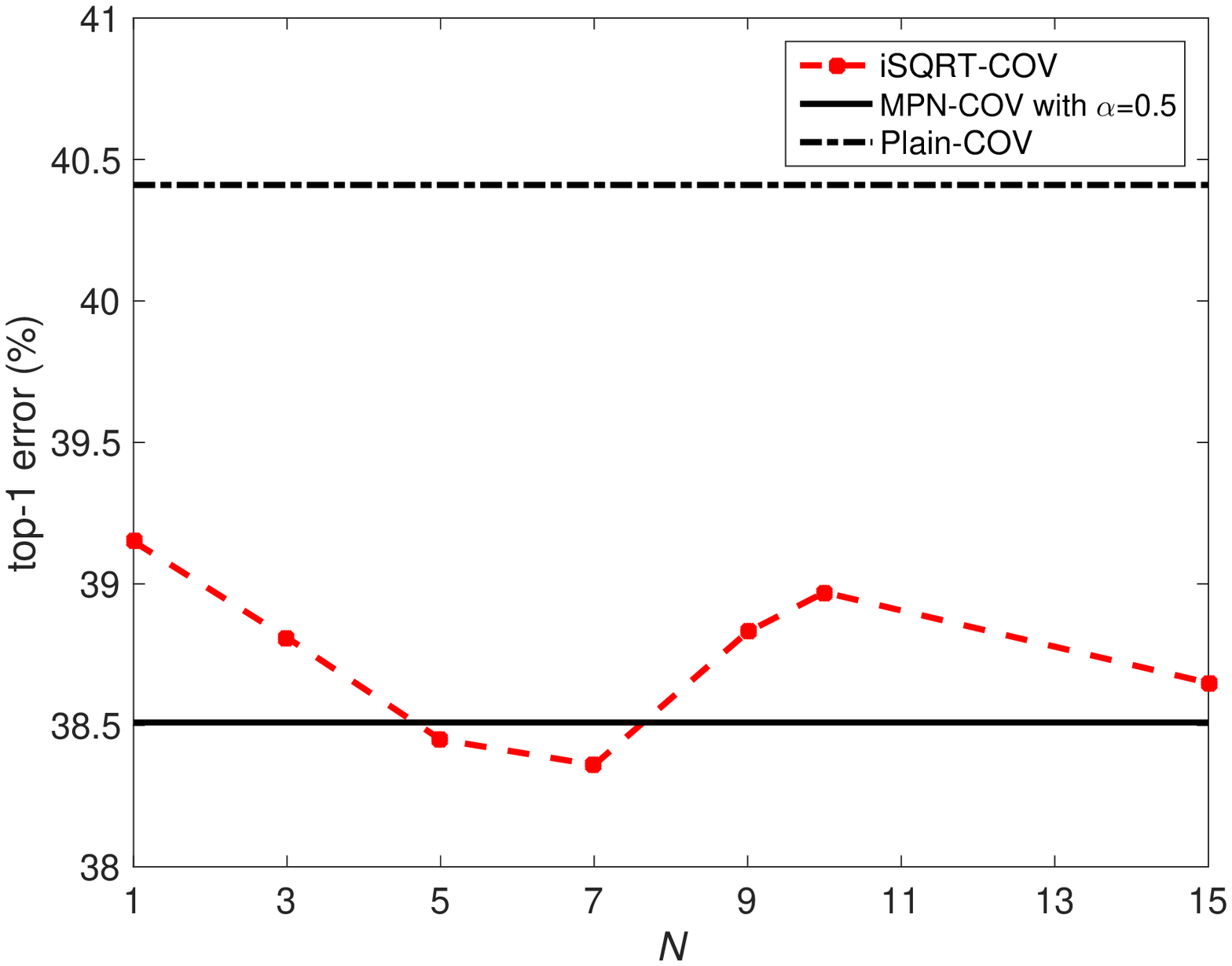}
		\label{fig:impactiterationN} 
	}
	\subfigure[]{
		\centering
		\includegraphics[width=0.31\textwidth]{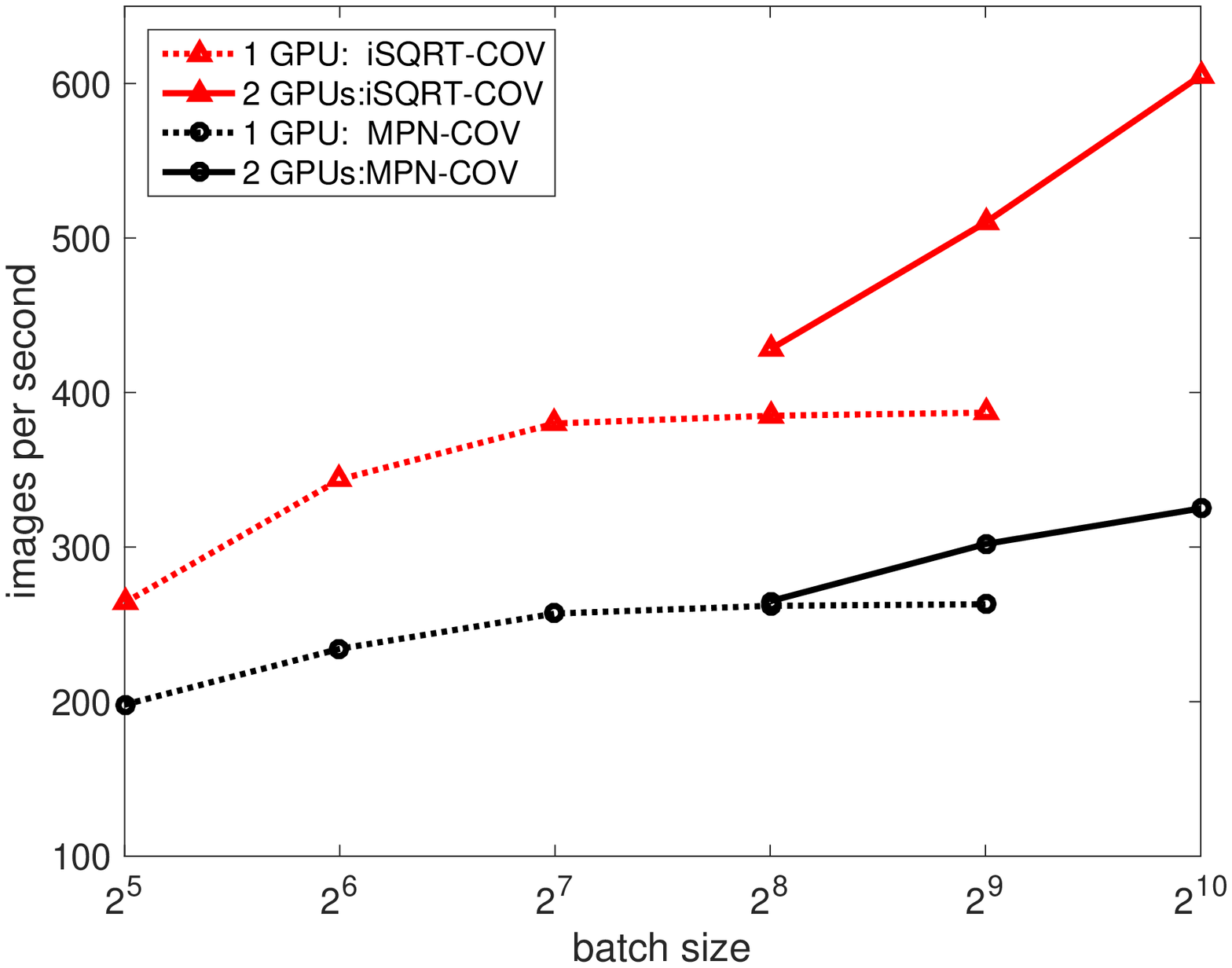} 
		\label{fig:Comparison_time}   
	}
	\caption{Ablation studies on ImageNet: (a) Effect of  $\alpha$  on MPN-COV with AlextNet, where Top-1 errors (1-crop prediction)  are reported and the bold line indicates the result of original AlexNet; (b) Impact of number $N$ of Newton-Schulz iterations on iSQRT-COV with AlexNet, evaluated with single-crop Top-1 error; (c) Images per second (FP$+$BP) of network training with AlexNet.}
\end{figure*}

\section{Experiments}
To evaluate the representation and generalization abilities of the proposed methods, we conduct experiments on both large-scale object classification and scene categorization, small-scale fine-grained visual recognition and texture classification as well as large-scale species classification competition. We first describe implementation details of our methods. Then we make ablation study of the proposed methods on large-scale ImageNet dataset, and compare with state-of-the-art methods on ImageNet \cite{imagenet_cvpr09} and Places365 \cite{zhou2017places}. Additionally, we verify the generalization ability of the proposed methods through transferring them to three fine-grained (i.e., Birds-CUB200-2011 \cite{WahCUB2002011}, Aircrafts \cite{aircraft} and Cars \cite{KrauseStarkDengFei-Fei_3DRR2013}) and two texture image datasets (i.e., Indoor67 \cite{QuattoniT09} and DTD \cite{cimpoi14describing}). Finally, we show the results on \href{https://www.kaggle.com/c/inaturalist-2018}{iNaturalist Challenge 2018} held in CVPR 2018 workshop on FGVC5.

\subsection{Implementation Details}\label{ipdts}
To implement MPN-COV block, we use the EIG algorithm on CPU in single-precision floating-point format, as the EIG algorithm on GPU is much slower, despite the overhead due to data transfer between GPU memory and CPU memory. Since MPN-COV allows non-negative eigenvalues, we truncate to zeros the eigenvalues smaller than $\mathrm{eps}(\lambda_{1})$, which is a matlab function denoting the positive distance from the largest eigenvalue $\lambda_{1}$ to its next larger floating-point number. For achieving the global Gaussian embedding block, as suggested in \cite{Ionescu_2015_ICCV}, we use SVD to compute square root SPD matrix because SVD is numerically more stable, and SVD algorithm is implemented on CPU as well. Meanwhile, we add a small positive number 1e-3 to the diagonal entries of Gaussian embedded matrices for numerical stability. The implementation of iSQRT-COV block is encapsulated in three computational modules, which accomplish forward\&backward computation of pre-normalization layer, Newton-Schulz iteration layer and post-compensation layer, respectively. We implement above methods based on the MatConvNet package \cite{vedaldi15matconvnet}; iSQRT-COV-Net is also implemented using Pytorch and TensorFlow. 

The proposed GCP blocks are inserted after the last convolution layer (with ReLU) of deep CNNs. Unless otherwise stated, we discard the last downsampling in the networks so that we  have larger number of features, and add one $1\times 1$ convolution with $d=256$ channels before our covariance pooling for all CNN architectures but AlexNet. As such, we have  $w\times h \times 256$ feature maps right before covariance pooling, where $w$ and $h$  are width and height of feature maps, obtaining  a $256\times 256$ covariance matrix (or 32,896-$d$ covariance representation after matrix vectorization). We train our networks using mini-batch stochastic gradient descent algorithm with a momentum of 0.9 and a weight decay of $5\times10^{-4}$. All programs run on two PCs each of which is equipped with a 4-core Intel i7-4790k@4.0GHz CPU, 32G RAM, 512GB Samsung PRO SSD and two Titan Xp GPUs. The remaining hyper-parameters of training networks will be described where appropriate in following subsections.

\subsection{Large-scale Object Classification on ImageNet}\label{ablation on imagenet}
\subsubsection{Ablation Study}\label{ablation}
We first analyze the proposed methods on large-scale ImageNet dataset, which contains 1.28M training images, 50K validation images and 100K testing images  collected from 1K classes. On this dataset, we follow \cite{Simonyan15} for data augmentation, and adopt the commonly used 1-crop or 10-crop prediction for performance evaluation. Following the common settings \cite{nips2012cnn,Simonyan15,He_2016_CVPR}, we report the results on the validation set. Specifically, we study the effect of parameter $\alpha$ on MPN-COV and compare various normalization methods on GCP, as well as assess impacts of number $N$ of Newton-Schulz iterations and post-compensation on iSQRT-COV, speedup ratio of training iSQRT-COV-Net over MPN-COV-Net, and effect of compact covariance representations. 

\textbf{MPN-COV: Effect of parameter $\alpha$} Our MPN-COV has a key parameter $\alpha$, and we evaluate it under AlexNet architecture. Our MPN-COV-Net is trained up to 20 epochs, where the learning rates follow exponential decay, changing from $10^{-1.2}$ to $10^{-5}$ with a batch size of 128. Fig.~\ref{fig:impact_alpha} shows top-1 error \textit{vs.} $\alpha$ using single-crop prediction. We first note that the \emph{Plain-COV} ($\alpha=1$, no normalization) produces an error rate of 40.41\%, about  $1.1\%$ less than the original AlexNet. When $\alpha<1$, the normalization function shrinks eigenvalues larger than 1.0 and stretches those less than 1.0. As $\alpha$ (less than 1.0) decreases, the error rate continuously gets smaller until the smallest value at around $\alpha=\frac{1}{2}$. With further decline of $\alpha$, however, we observe the error rate grows consistently and soon is larger than that of the Plain-COV. Note that over the interval $[0.4,0.9]$ the performance of MPN-COV varies insignificantly. When $\alpha>1$, the effect of normalization is contrary, i.e., eigenvalues less than 1.0 are shrunk while those larger than 1.0 are stretched, which is not beneficial for covariance representations as indicated by the consistent growth of the error rates. We note that similar trend of effect of matrix power on classification performance was also reported in~\cite{lin2017improved}. As such, we set $\alpha=\frac{1}{2}$ throughout the following experiments.

\textbf{MPN-COV: Comparison of various normalization methods} We compare four kinds of normalization methods, i.e., MPN, M-Fro, M-$\ell_{2}$ and element-wise power normalization followed by $\ell_{2}$-normalization (E-PN for short)~\cite{sanchez,lin2015bilinear}. Table~\ref{table:different-norm} gives the comparison results, where we can see that all GCP methods except MPN+E-PN outperform the original network, and all normalization methods improve over the plain COV; among them, our MPN obtains the best result, and outperforms M-Fro, M-$\ell_{2}$ and E-PN by ${\sim}1.3\%$, ${\sim}1.1\%$ and ${\sim}1.3\%$, respectively. Note that GCP with MPN (i.e., MPN-COV) followed by further normalization, either by M-Fro, M-$\ell_{2}$ or E-PN, produces negative gains, so we do not perform any further normalization on our MPN-COV in following experiments.

\begin{table}[tb]
	\footnotesize
	\centering
	\caption{Top-1  error (\%, 1-crop prediction) of GCP with various normalization methods with AlexNet on ImageNet.}
	\label{table:different-norm}
	\begin{tabular}{lllllc}
		\hline
		method & MPN (ours)  & M-Fro & M-$\ell_2$ & E-PN   & top-1 Err. \\
		\hline
		Baseline  & -- & -- & -- & -- & 41.52\\
		\hline
		\multirow{7}{*}{GCP}
		& $\times$    & $\times$    & $\times$   &  $\times$   &   40.41 \\
		& $\checkmark$   & $\times$  & $\times$   &  $\times$   &   \textbf{38.51}  \\ 
		& $\checkmark$   & $\checkmark$   & $\times$   &  $\times$   &  39.93\\
		& $\checkmark$   & $\times$    & $\checkmark$  &  $\times$   &  39.62\\
		&  $\checkmark$   & $\times$    &  $\times$ &  $\checkmark$   &  40.75\\
		& $\times$    & $\checkmark$   & $\times$   &  $\times$   &   39.87  \\
		& $\times$    & $\times$    & $\checkmark$  &  $\times$   &  39.65\\
		& $\times$    & $\times$    & $\times$  &  $\checkmark$   &  39.89\\
		\hline
	\end{tabular}
\end{table}

\textbf{iSQRT-COV: Number $N$ of Newton-Schulz iterations} Here we assess impact of $N$ by employing AlexNet as a backbone model, and using the same hyper-parameters as MPN-COV except the initial learning rate, which is set to $10^{-1.1}$. The top-1 error rate (1-crop prediction) \textit{vs.} $N$ is illustrated in Fig.~\ref{fig:impactiterationN}. With single one iteration, our iSQRT-COV outperforms Plain-COV by $1.3\%$. As iteration number grows, the error rate of iSQRT-COV gradually declines. With 3 iterations, iSQRT-COV is comparable to MPN-COV, having only 0.3\% higher error rate, while performing marginally better than MPN-COV between 5 and 7 iterations. The results after $N=7$ show growth of iteration number is not helpful for decreasing error. As larger $N$ incurs higher computational cost, we set $N$ to 5 in the remaining experiments for balancing efficiency and effectiveness. The similar trend was also shown in \cite{lin2017improved}. An interesting phenomenon is that approximate matrix square root normalization in iSQRT-COV achieves a little gain over the exact one obtained via EIG, which may suggest the fact that matrix square root is not the optimal normalization method for covariance pooling, encouraging development of better normalization methods in the future work.

\begin{table}[t]
	\footnotesize
	\centering
	\caption{Impact of post-compensation on iSQRT-COV  with ResNet-50 on ImageNet, evaluated with 1-crop prediction.}
	\label{table:Post-compensation}
	\begin{tabular}{c|l|c|c}
		\hline
		Pre-normalization & Post-compensation & Top-1 Err.   &  Top-5 Err. \\
		\hline
		\multirow{3}{*}{Trace} & $\quad\;$w/o       &   N/A    & N/A     \\
		& $\quad\;$w/~~BN~\cite{icml2015_ioffe15}       &  23.12     & 6.60     \\
		& $\quad\;$w/~~Trace        &  $\textbf{22.14}$ & $\textbf{6.22}$ \\
		\hline
	\end{tabular}
\end{table}

\begin{table}[thb]
	\footnotesize
	\centering
	\caption{Error rate (\%, 1-crop  prediction) and extra network parameters (Params.) of iSQRT-COV using various compact covariance representations (Repr.) on ImageNet. ResNet-50 is used as backbone model and is compared as baseline. }
	\label{table:compact-iSQRT-COV}
	\begin{tabular}{l|c|c|c|l}
		\hline
		Method  & DR & Repr. & Top-1/Top-5 & Params.  \\
		\hline
		GAP \cite{He_2016_CVPR}  & N/A & 2K& 24.7/7.8 & 25.56M  \\
		\hline
		\multirow{8}{*}{ \tabincell{c}{iSQRT\\-COV}}  &  $2048\rightarrow256$  & 32K       & 22.14/6.22& +30.53M \\
		&  $2048\xrightarrow{512}256$ & 32K        & 21.72/5.99 & +31.36M \\
		&  $2048\rightarrow128$  & 8K        & 22.78/6.43 & +6.26M \\
		&  $2048\xrightarrow{512}128$ & 8K       & 22.33/6.28 & +7.07M \\
		&  $2048\rightarrow64$   & 2K    & 23.73/6.99  & +0.13M \\
		&  $2048\xrightarrow{256}64$ & 2K        & 22.98/6.61  & +0.54M \\
		&  $4th$ row \& $1G$ & 2K       & 22.40/6.35 & +18.9M\\
		&  $4th$ row \& $2G$ & 2K       &  22.84/6.60 & +9.6M\\
		&  $4th$ row \& 4G & 2K        & 23.11/6.77& +4.3M\\
		\hline
	\end{tabular}
\end{table}

\textbf{iSQRT-COV: Significance of post-compensation} As discussed in Section~\ref{iSQRT-COV}, post-compensation in iSQRT-COV helps to eliminate the side effect resulting from pre-normalization. We verify its significance with ResNet-50 architecture. Table~\ref{table:Post-compensation} summarizes impact of different schemes on iSQRT-COV-Net, including simply do nothing (i.e., without post-compensation), Batch Normalization (BN)~\cite{icml2015_ioffe15} and our post-compensation scheme. We note that iSQRT-COV-Net fails to converge without post-compensation. Careful observations show that in this case  the gradients are very small (on the order of $10^{-5}$), and larger learning rate helps little. Option of BN helps the network converge, but producing about 1\% higher top-1  error rate than our post-compensation scheme. The comparison above suggests that our post-compensation is essential for achieving state-of-the-art results with deep CNN architectures, e.g., ResNet-50.

\textbf{MPN-COV \emph{VS.} iSQRT-COV in training speed} To show acceleration effect of iSQRT-COV, we compare in Fig.~\ref{fig:Comparison_time} training speed between MPN-COV-Net and iSQRT-COV-Net with both 1-GPU and 2-GPU configurations. For 1-GPU configuration, the speed gap \emph{vs.} batch size between the two methods keeps nearly constant. For 2-GPU configuration, their speed gap becomes more significant when batch size gets larger. As can be seen, the speed of iSQRT-COV-Net continuously grows with increase of batch size while MPN-COV-Net tends to saturate when batch size is larger than 512. Clearly, by avoiding GPU unfriendly EIG or SVD, iSQRT-COV can speed up MPN-COV and make better use of computing power of multiple GPUs.

\textbf{Compact covariance representations} In default setting, our methods output a 32k-dimensional covariance representation. In this paper, the strategies of progressive $1\times1$ convolutions and group convolution \cite{XieGDTH17} are proposed to further compress covariance representations. We evaluate them using iSQRT-COV with ResNet-50. Table~\ref{table:compact-iSQRT-COV} summarizes results of our iSQRT-COV and the extra increased parameters (Params.) with respect to ResNet-50 under various compact settings. Specifically, we use a single $1 \times 1$ convolution to decrease dimension of convolution features from 2048 to 256, 128 and 64. Then, progressive dimensionality reduction (DR) is performed based on two consecutive $1\times1$ convolutions, i.e., $2048\rightarrow512\rightarrow256$, $2048\rightarrow512\rightarrow128$ and $2048\rightarrow256\rightarrow64$. We do not employ more $1 \times 1$ convolutions as they will bring additional parameters. Meanwhile, size of covariance representation with $2048\rightarrow512\rightarrow128$ ($4th$ row) is further reduced to 2K using group convolution of one group, two and four groups, which are indicated by $1G$, $2G$ and $4G$, respectively.

As listed in Table~\ref{table:compact-iSQRT-COV}, by using a single $1 \times 1$ convolution, the recognition error increases about $1.0\%$ and $1.6\%$ when $d$ decreases from 256 to 128 and 64, respectively. The results of progressive DR versions are better than their counterparts based on a single $1 \times 1$ convolution, showing the strategy of progressive DR is helpful to preserve classification accuracy for small-size representations. Combining progressive $1 \times 1$ convolutions with group convolution, we can obtain $22.4\%$ top-1 error with 2$\text{K}$-dimensional representations, still clearly outperforming the original ResNet-50 with first-order GAP. Above results verify the effectiveness of our compact covariance representations. Our compact strategies can obtain effective small-size representations, very suitable for large scale species classification problem, which will be shown in Section~\ref{iNaturalist2018}.

\begin{table}[t]
	\footnotesize
	\centering
	\caption{Error rate (\%, 1-crop  prediction) and time of FP+BP (ms) per image of different covariance pooling methods with AlexNet on ImageNet. Numbers in parentheses indicate FP time.  $^{*}$Following~\cite{lin2017improved}, improved B-CNN  successively  performs matrix square root and E-PN.}
	\label{table:second-order-AlexNet}
		\centering
		\begin{tabular}{l|c|c|c}
			\hline
			Method & Top-1 Err.   &  Top-5 Err.  & Time \\
			\hline
			AlexNet~\cite{nips2012cnn}       &  41.8     & 19.2  &  1.32 (0.77) \\
			\hline
			B-CNN~\cite{lin2015bilinear}          &  39.89 & 18.32 & 1.92 (0.83)\\
			DeepO$_{2}$P~\cite{Ionescu_2015_ICCV} &  42.16 & 19.62  & 11.23 (7.04)\\
			Improved B-CNN$^{*}$\cite{lin2017improved}     & 40.75         &  18.91  &  15.48 (13.04)  \\
			\hline
			MPN-COV-Net                           &  38.51 & 17.60 & 3.89 (2.59)\\
			G$^{2}$DeNet      &    38.48      & 17.59     &   9.86 (5.88) \\
			iSQRT-COV-Net(Frob.)   &  38.78 &  17.67  & 2.56 (0.81)\\
			iSQRT-COV-Net(trace) &  \textbf{38.45} &  \textbf{17.52} & 2.55 (0.81) \\
			\hline
		\end{tabular}
\end{table}


\begin{figure}[tb]
	\centering
	\begin{minipage}[b]{0.45\linewidth}
		\centering
		\includegraphics[width=1.0\textwidth]{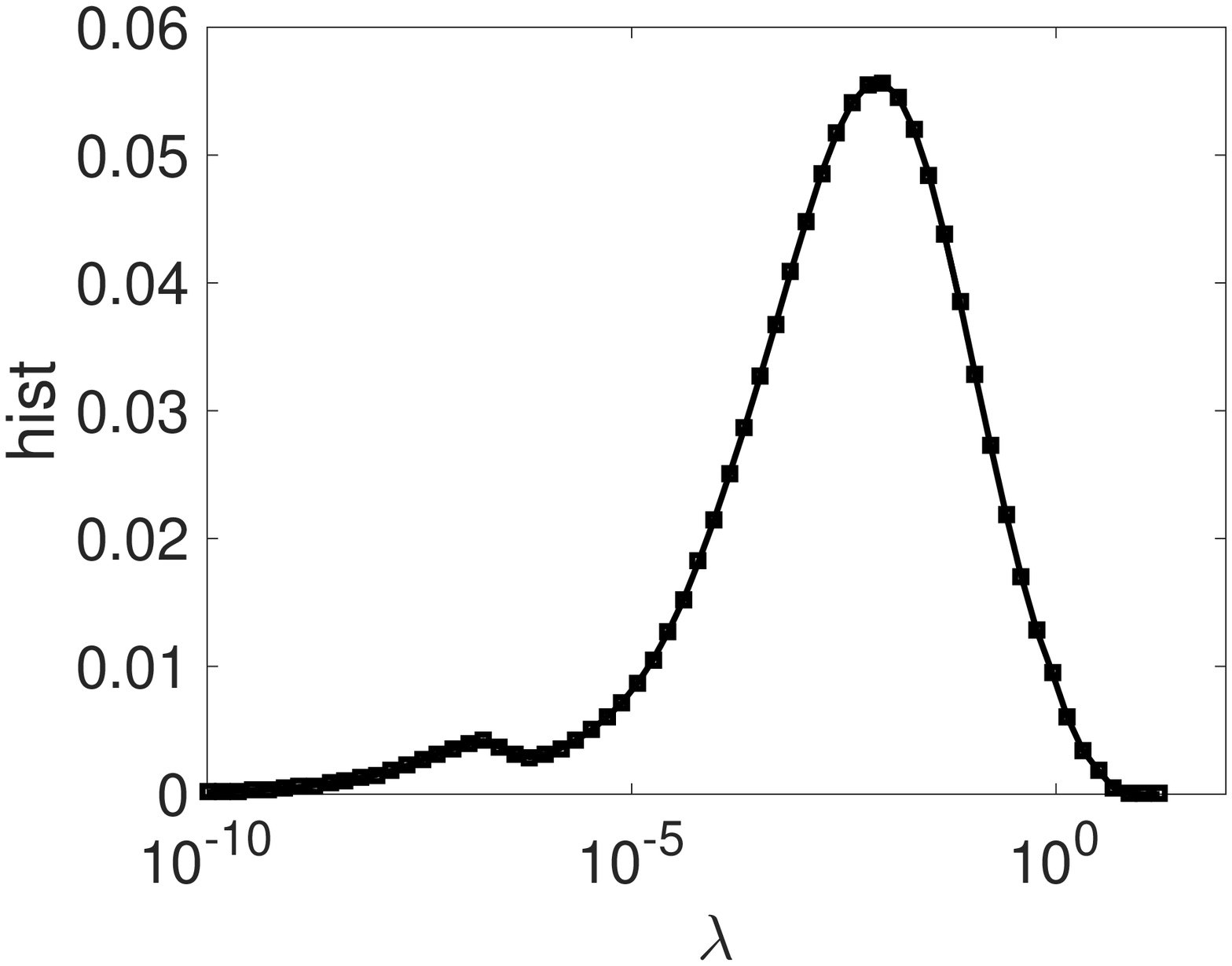}
	\end{minipage}
	\begin{minipage}[b]{0.45\linewidth}
		\centering
		\includegraphics[width=1.0\textwidth]{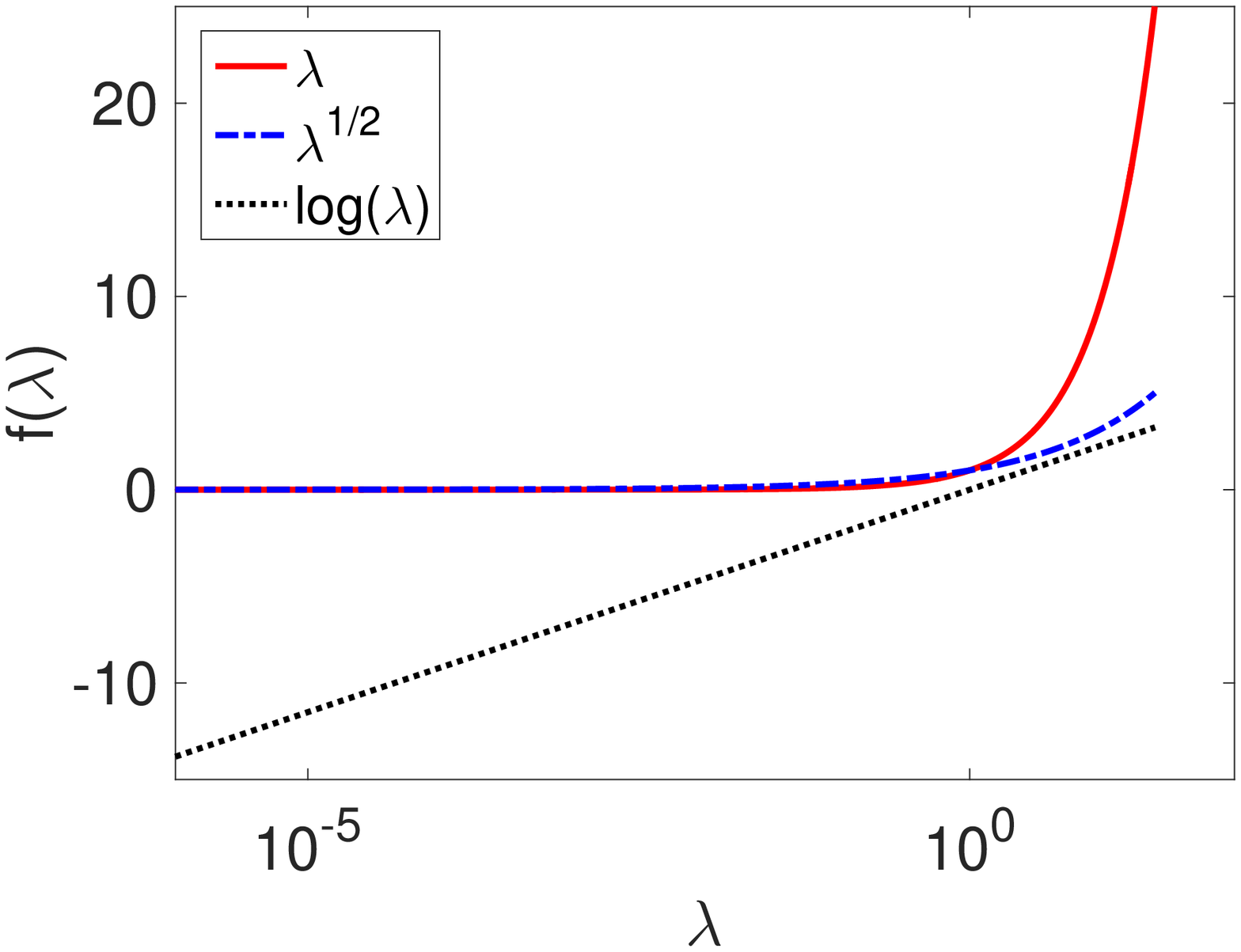}
	\end{minipage}
	\caption{Illustration of empirical distribution of eigenvalues (left) and  normalization functions (right).}
	\label{figure:hist-and-norm-functions}
\end{figure}

\begin{figure*}
	\centering
	\subfigure[]{
		\centering
		\includegraphics[width=0.3\textwidth]{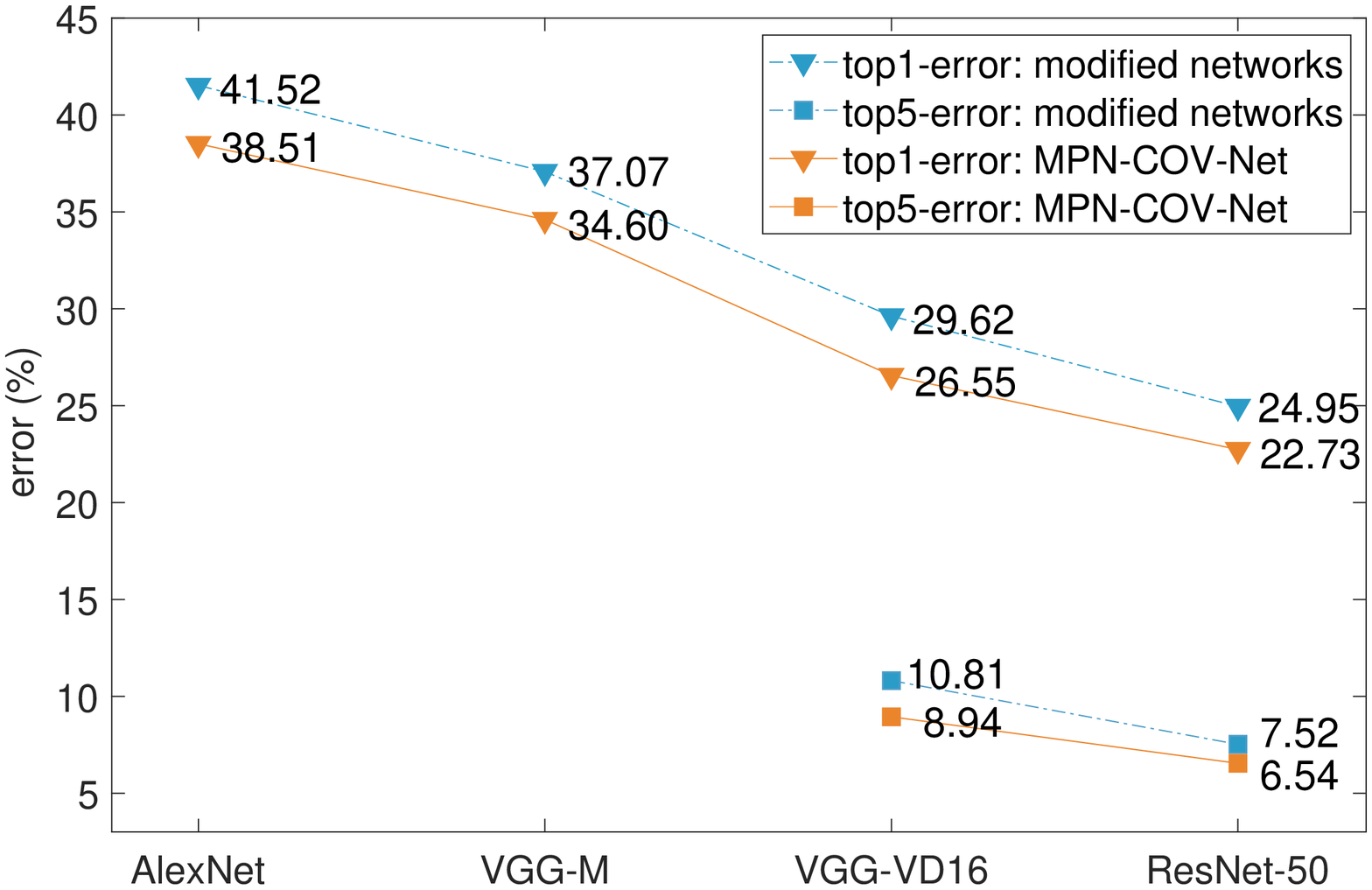} 
		\label{fig:modified} 
	}
	\subfigure[]{
		\centering
		\includegraphics[width=0.3\textwidth]{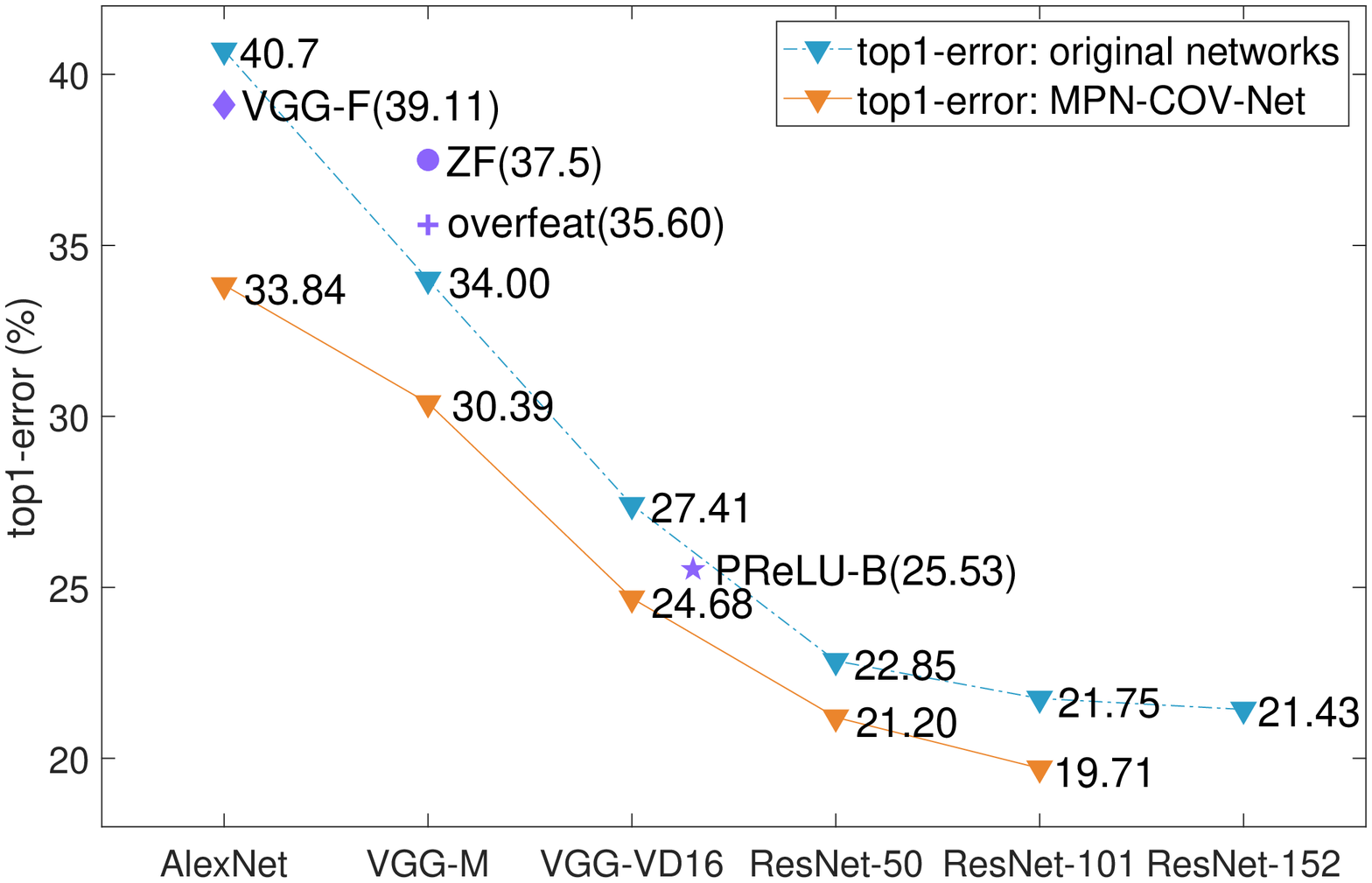}
		\label{fig:top1} 
	}
	\subfigure[]{
		\centering
		\includegraphics[width=0.3\textwidth]{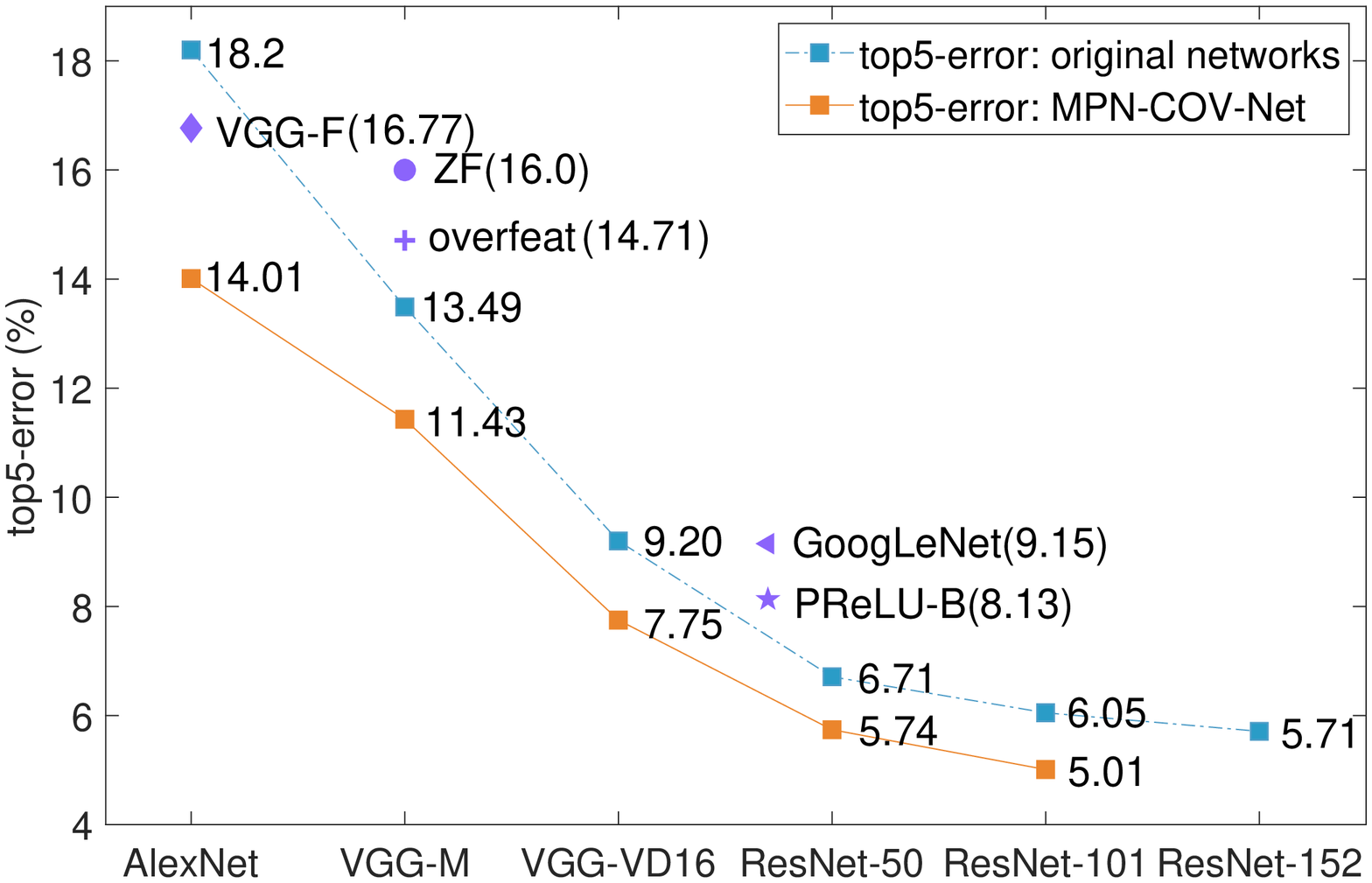} 
		\label{fig:top5}
	}
	\caption{Comparison of classification error  with different CNN models. (a) show MPN-COV-Net \emph{vs.} corresponding modified first-order networks evaluated with single-crop Top-1/Top-5 errors. (b) and (c) show MPN-COV-Net \emph{vs.} original  first-order networks and \emph{vs.} state-of-the-arts evaluated with ten-crop Top-1 error and Top-5 error, respectively.}
\end{figure*}

\subsubsection{Comparisons with Different Methods}\label{Comparisons}
\textbf{Comparison with other deep global GCP methods} We compare different variants of MPN-COV with existing deep global GCP methods using AlexNet architecture, including B-CNN \cite{lin2015bilinear}, DeepO$_{2}$P \cite{Ionescu_2015_ICCV} and improved B-CNN \cite{lin2017improved}. All methods are trained from scratch on ImageNet. We implement B-CNN, DeepO$_{2}$P and improved B-CNN using public available source code released by the respective authors, and try our best to tune hyper-parameters for them. Note that we use the suggested implementation of improved B-CNN in \cite{lin2017improved}, i.e., FP by SVD and BP by Lyapunov equation. The results of different methods are listed in Table~\ref{table:second-order-AlexNet}. We can see that our MPN-COV-Net,  G$^{2}$DeNet and iSQRT-COV-Net obtain similar results, and clearly outperform B-CNN, DeepO$_{2}$P and improved B-CNN. We owe the gains to matrix power normalization, benefiting robust covariance estimation and proper usage of geometry. The improved B-CNN achieves an unsatisfactory result, which can be regraded as performing element-wise power normalization followed by $\ell_{2}$ normalization behind our MPN-COV. This result suggests that further normalization hurts MPN-COV on large-scale ImageNet classification, which is consistent with the observation in Table~\ref{table:different-norm}. Additionally, for iSQRT-COV-Net, pre-normalization by trace performs better than Frobenius norm. G$^{2}$DeNet combining mean vector achieves moderate gains over MPN-COV-Net using larger-size representations, and so we mainly report the results of MPN-COV-Net and iSQRT-COV-Net (trace) in following comparisons.

\textbf{Why LERM does not work well?} DeepO$_{2}$P  exploits matrix logarithm normalization (i.e., LERM \cite{Arsigny2005}) to exploit geometry of covariance, and we claim it is not suitable for deep CNNs. Firstly,  the Pow-E metric in MPN-COV improves numerical stability of covariance matrices over LERM. The LERM requires the eigenvalues involved to be \emph{strictly positive}~\cite{Arsigny2005} while the Pow-E metric allows \emph{non-negative} eigenvalues~\cite{Dryden2009}. For LERM   the  common method is to add a small positive number $\epsilon$  to  eigenvalues  for numerical stability. Although  $\epsilon$ can be decided by cross-validation, it is difficult to seek the optimal $\epsilon$ suitable for training of deep CNN. For example, \cite{Ionescu_2015_ICCV} suggest $\epsilon=10^{-3}$, which will  smooth out eigenvalues less than $10^{-3}$. In contrast, the Pow-E metric do not need such a remedy. Furthermore,  from the distribution perspective, the matrix logarithm reverses the order of the significance of eigenvalues, harmful for covariance representations. To make a qualitative analysis, we randomly select 300,000 images from training set of ImageNet, and estimate per-image sample covariances using the outputs of the last convolution layer in AlexNet model. Then, we compute eigenvalues of all covariances using EIG in single-precision floating-point format. The histogram of eigenvalues is shown in Fig.~\ref{figure:hist-and-norm-functions} (left), where zero eigenvalues are excluded for better view.  Fig.~\ref{figure:hist-and-norm-functions} (right) shows the two normalization functions  over $[10^{-5}, 10]$. We can see $\log(\lambda)$ considerably changes the eigenvalue magnitudes, reversing the order of significance of eigenvalues, e.g., a significant eigenvalue $\lambda=50\mapsto \log(\lambda)\approx 3.9$ but an insignificant one $\lambda=10^{-3}\mapsto \log(\lambda)\approx-6.9$. Since significant eigenvalues are generally more important in that they capture the statistics of principal directions along which the feature variances are larger, matrix logarithm will harm covariance pooling.

\begin{table}[thb]
	\footnotesize
	\centering
	\caption{Comparison of classification error  (\%, 1-crop prediction)   with local second-order networks with  ResNet as backbone models on ImageNet.}
	\label{table:ImageNet-ResNet}
	\begin{tabular}{l|c|c|c}
		\hline
		Method  & Backbone model &Top-1 Err.   &  Top-5 Err. \\
		\hline
		ResNet-50~\cite{He_2016_CVPR}  & \multirow{7}{*}{ResNet-50} & 24.7 &  7.8 \\
		FBN~\cite{LiYanghao_2017_ICCV}  & &24.0 & 7.1 \\
		SORT~\cite{Wang_2017_ICCV}    & &23.82 & 6.72  \\
		SE-Net~\cite{SENet18}   &     & 23.29 &  6.62 \\
		CBAM~\cite{Woo_2018_ECCV}   &     & 22.66 &  6.31 \\
		MPN-COV-Net                       & &22.73  & 6.54 \\
		iSQRT-COV-Net   &     & \textbf{22.14}&  \textbf{6.22} \\
		\hline
		ResNet-101~\cite{He_2016_CVPR}  & \multirow{4}{*}{ResNet-101} & 23.6 &  7.1 \\
		SE-Net~\cite{SENet18}  &     & 22.38 &  6.07 \\
		CBAM~\cite{Woo_2018_ECCV}   &     & 21.51 &  5.69 \\
		iSQRT-COV-Net   &     & \textbf{21.21} &  \textbf{5.68} \\
		\hline
		ResNet-152~\cite{He_2016_CVPR}  & ResNet-152 & 23.0 &  6.7 \\
		\hline
		DenseNet-201~\cite{Huang_2017_CVPR}  & \multirow{2}{*}{DenseNet-201} & 22.58 &  6.34 \\
		iSQRT-COV-Net   &     & \textbf{20.69} &  \textbf{5.48} \\
		\hline
	\end{tabular}
\end{table}

\begin{table*}
	\footnotesize
	\centering
	\caption{Comparison of classification error (\%, 10-crop prediction) on Places365 dataset. B-CNN and iSQRT-COV-Net use ResNet-50 as backbone model.}
	\label{tab:Place365}
	\begin{tabular}{lccccccc}
		\hline
		&  VGG-VD16 \cite{Simonyan15} & GoogleNet \cite{Szegedy_2015_CVPR}  &  ResNet-50 \cite{He_2016_CVPR}  & ResNet-152 \cite{zhou2017places} & B-CNN  &  iSQRT-COV-Net & iSQRT-COV + E-PN \\
		\hline
		Top-1 Error  & 44.76 &  46.37 & 44.82 & 45.26 & 44.24 & \textbf{43.68} & 45.34  \\
		Top-5 Error & 15.09 & 16.12  & 14.71 & 14.92 & 14.27  & \textbf{13.73} & 15.13 \\
		\hline
	\end{tabular}
\end{table*}

\textbf{Comparison with deep local second-order networks} In Table~\ref{table:ImageNet-ResNet}, we compare our MPN-COV-Net and iSQRT-COV-Net, using ResNet-50 architecture, with deep local second-order networks \cite{LiYanghao_2017_ICCV,Wang_2017_ICCV}. Clearly, the two networks integrating local second-order statistics improve over the original one. Our MPN-COV-Net and iSQRT-COV-Net employing matrix square root normalization, are superior to  FBN~\cite{LiYanghao_2017_ICCV} and SORT~\cite{Wang_2017_ICCV}. These results demonstrate again the effectiveness of our MPN-COV. Similar to the results with AlexNet, iSQRT-COV-Net outperforms MPN-COV-Net by 0.6\% in top-1 error. 

\textbf{Comparison with various CNN architectures} In the end of this subsection, we evaluate our MPN-COV with different CNN architectures. Since we modify the existing CNN architectures (as described in Section~\ref{ipdts}), we first compare our MPN-COV with the modified CNN architectures. Fig.~\ref{fig:modified} illustrates the results of different networks with 1-crop prediction, from it we can see that our MPN-COV outperforms consistently the corresponding modified CNN architectures by a clear margin. Then, we combine the proposed MPN-COV with five deep CNN models, including AlexNet~\cite{nips2012cnn}, VGG-M~\cite{ChatfieldSVZ14}, VGG-VD16~\cite{Simonyan15}, ResNet-50~\cite{He_2016_CVPR} and ResNet-101~\cite{He_2016_CVPR}. Here we compare MPN-COV-Net with the original CNN models and the networks with similar architectures.  The top-1 error and top-5 error of different networks with 10-crop prediction are given in Fig.~\ref{fig:top1} and Fig.~\ref{fig:top5}, respectively. The results of compared methods are duplicated from the original papers.  

Using AlexNet as backbone model, we compare MPN-COV-Net with the original AlexNet and VGG-F~\cite{ChatfieldSVZ14}.  Our MPN-COV-Net performs much better than both of them. Comparing MPN-COV-Net using VGG-M~\cite{ChatfieldSVZ14} with the original one,  Zeiler \& Fergus~\cite{ZeilerF14} and OverFeat \cite{SermanetEZMFL14}, our MPN-COV-Net shows much better performance than them. Using VGG-VD16 as backbone model, our MPN-COV-Net outperforms the original VGG-VD16~\cite{Simonyan15} by ${\sim}2.7\%$ in terms of top-1 error, and performs better than GoogleNet~\cite{Szegedy_2015_CVPR}  and PReLU-net B~\cite{HeZRS15}  by ${\sim}1.4\%$  and ${\sim}0.4\%$ in terms of top-5 error, respectively. For ResNet-50 and ResNet-101 architectures, the results show that our MPN-COV-Net performs $1.65\%$ and $2.04\%$ better than the original ones in terms of top-1 error, respectively. Note that our MPN-COV with ResNet-50 and ResNet-101 outperform the original ResNet-101 and ResNet-152 based on first-order GAP, respectively. We also compare iSQRT-COV-Net, in Table~\ref{table:ImageNet-ResNet}, with the recently proposed SE-Net~\cite{SENet18}, CBAM~\cite{Woo_2018_ECCV} using ResNet-50 and ResNet-101 in 1-crop prediction. Our iSQRT-COV-Net consistently improves SE-Net and CBAM. Compared with DenseNet-201~\cite{Huang_2017_CVPR}, our iSQRT-COV-Net obtains ${\sim}1.9\%$ gains in top-1 error. Additionally, we make a comprehensive comparison of iSQRT-COV with GAP using ResNets in terms of efficiency and effectiveness, and the corresponding results can be found in Appendix III. These results show our GCP methods can effectively improve CNNs with various architectures with affordable model complexity. 

\subsection{Large-scale Scene Categorization on Places365}
We evaluate our methods on large-scale scene categorization using Places365 dataset \cite{zhou2017places}, which contains about 1.8 million training images and 36,500 validation images collected from 365 scene categories. Following the common settings in \cite{zhou2017places}, we resize all images to $256\times256$ and randomly crop a $224\times224$ image patch or its flip for training. The inference is performed with 10-crop prediction, and we report the results on validation set for comparison. Here we compare our iSQRT-COV-Net, using ResNet-50, with five kinds of CNN models, i.e., GoogleNet \cite{Szegedy_2015_CVPR}, VGG-VD16 \cite{Simonyan15}, ResNet-50 \cite{He_2016_CVPR}, ResNet-152 \cite{He_2016_CVPR} and B-CNN with ResNet-50. The results of different methods are given in Table~\ref{tab:Place365}, from it we can see that our iSQRT-COV-Net achieves the best results, and outperforms the original ResNet-50 by about $1.2\%$ and $1\%$ in Top-1 and Top-5 errors, respectively. Meanwhile, iSQRT-COV-Net is superior to B-CNN under the same settings. Furthermore,  E-PN after our iSQRT-COV block  is hurtful in the setting of training from scratch on Places365, which is similar to the results on ImageNet as in Table~\ref{table:different-norm}.  The results on both ImageNet and Places365 verify our proposed methods can significantly improve the representation ability of deep CNNs, achieving much lower classification error. 

\begin{table}[thb]
	\caption{Comparison of classification accuracy (\%) with state-of-the-art methods on fine-grained benchmarks.}
	\label{table:Fine-grained}
	\footnotesize
	\centering
	\begin{tabular}{c|c|c|c|c}
		\hline
		& Method   & Birds   &  Aircrafts  & Cars \\
		\hline 
		\multirow{13}{*}{ \rotatebox{90}{{VGG-D16}}}
		& VGG-VD16~\cite{lin2015bilinear}                            & 70.4  & 76.6      &  79.8\\
		& NetVLAD~\cite{Arandjelovic_2016_CVPR}                            & 81.9  & 81.8      &  88.6\\
		& NetFV~\cite{pami/LinRM18}                            & 79.9  & 79.0      &  86.2\\
		& B-CNN~\cite{lin2015bilinear}                             & 84.0  & 83.9      &  90.6\\
		& CBP~\cite{Gao_2016_CVPR}                             & 84.3  & 84.1      &  91.2\\
		& LRBP~\cite{Kong_Charless_2017_CVPR}                            & 84.2  & 87.3      &  90.9\\
		& KP~\cite{Cui_2017_CVPR}                             & 86.2  & 86.9      &  92.4\\
		& HIHCA~\cite{Cai_2017_ICCV}                         & 85.3  & 88.3      &  91.7 \\
		& Improved B-CNN\cite{lin2017improved}     & 85.8  & 88.5      &  92.0\\
		& SMSO~\cite{DBLP:conf/eccv/YuS18}     & 85.01  & N/A      &  N/A\\
		\cline{2-5}
		& MPN-COV-Net                & 86.7  & 89.9      &  92.2\\
		& G$^2$DeNet                & 87.1  & 89.0      &  92.5\\
		& iSQRT-COV-Net                                       &               \textbf{87.2}  & \textbf{90.0}      &  \textbf{92.5}\\
		\hline
		\multirow{5}{*}{ \rotatebox{90}{{ResNet-50}}}
		& CBP~\cite{Gao_2016_CVPR}                                & 81.6    & 81.6      &  88.6\\
		& KP~\cite{Cui_2017_CVPR}                                 & 84.7    & 85.7      &  91.1\\
		& SMSO~\cite{DBLP:conf/eccv/YuS18}   & 85.77  & N/A      &  N/A \\
		\cline{2-5}
		& iSQRT-COV-Net (8K)    & 87.3    & 89.5      &  91.7\\
		& iSQRT-COV-Net (32K)  & \textbf{88.1}  & \textbf{90.0}  & \textbf{92.8}\\
		\hline
		\multicolumn{2}{l|}{iSQRT-COV-Net with ResNet-101}       & \textbf{88.7}  & \textbf{91.4}      &  \textbf{93.3}  \\
		\hline
	\end{tabular}
\end{table}

\subsection{Experiments on Small-scale Datasets}
In this section, we evaluate our methods on small-scale fine-grained visual recognition and texture classification tasks under the setting of fine-tuning, and assess effect of different fine-tuning strategies.
\subsubsection{Fine-grained Visual Recognition}
To assess the generalization (transfer) ability of the proposed methods, we pre-train the networks with our GCP on ImageNet and evaluate their performance on fine-grained visual recognition task by fine-tuning. The experiments are conducted on three benchmarks. Among them, \emph{Birds-CUB200-2011} \cite{WahCUB2002011} is a challenging dataset, including 11,788 images from 200 bird species. \emph{FGVC-aircraft} \cite{aircraft} is a part of the FGComp 2013 challenge, which consists of 10,000 images across 100 aircraft classes. \emph{FGVC-Cars} \cite{KrauseStarkDengFei-Fei_3DRR2013} is also presented as a part of the FGComp 2013 challenge, containing 16,185 images from 196 car categories. We employ the fixed training/testing splits provided by the dataset developers, and train or evaluate our networks using neither part annotations nor bounding boxes. For fair comparison, we follow~\cite{lin2015bilinear} for experimental setting and evaluation protocol. Specifically, we resize the shorter side of input images to 448, and crop center $448\times 448$ patches. We replace  1000-way softmax layer of our pre-trained networks by a k-way softmax layer, where $k$ is number of classes in the corresponding fine-grained dataset, and fine-tune the networks for 50$\sim$100 epochs with a small learning rate $lr$ (e.g., $10^{-2.1}$) for all layers except the last FC layer, which is set to $5\times lr$. The random horizontal flipping is used for data augmentation. After fine-tuning, we perform  $\ell_{2}-$normalization on the outputs of our GCP blocks, and feed them to train $k$ one-vs-all linear SVMs with parameter $C=1$. We predict the label of a test image by averaging SVM scores of the image and its horizontal flip.

\begin{table}[t]
	\footnotesize
	\centering
	\caption{Comparison of classification accuracy ($\%$) with state-of-the-art methods  on DTD and Indoor67.}
	\label{table:texture}
	\centering
	\begin{tabular}{l|c|c|c}
		\hline
		Method & Backbone Model   &  DTD  & Indoor67 \\
		\hline
		VGG-VD16 \cite{Dai_2017_CVPR}        &  VGG-VD16     & 62.9$\pm$0.8  &  67.6 \\
		\hline
		B-CNN \cite{lin2015bilinear}        &  VGG-VD16     & 72.9$\pm$0.8  &  79.0 \\
		\hline
		FASON  \cite{Dai_2017_CVPR}        & VGG-VD16 & 72.9$\pm$0.7 & 80.8\\
		\hline
		Deep-TEN \cite{ZhangXD17} &  ResNet-50 & N/A  & 76.2 \\
		\hline
		MFAFV-Net \cite{DixitV16} & VGG-VD16         &  N/A  &  81.1  \\
		\hline
		\multirow{2}{*}{SMSO~\cite{DBLP:conf/eccv/YuS18}}	     & VGG-VD16        &  69.26  &  79.45  \\
		& ResNet-50         & 72.51   &  79.68\\
		\hline
		\multirow{2}{*}{iSQRT-COV-Net}	     & VGG-VD16         &  \textbf{74.0$\pm$0.8}  &  \textbf{81.4}  \\
		& ResNet-50         & \textbf{74.8$\pm$1.0}   &  \textbf{83.5}\\
		\hline
	\end{tabular}
\end{table}

Table~\ref{table:Fine-grained} presents classification results of different methods, where our networks significantly improve the original networks under either VGG-VD16 or ResNet-50. When VGG-VD16 is used as backbone model, our proposed methods are superior to both deep CNNs with trainable BoVW (i.e.,  NetVLAD~\cite{Arandjelovic_2016_CVPR} and NetFV~\cite{pami/LinRM18}) and deep CNNs with global approximate high-order pooling (i.e., CBP~\cite{Gao_2016_CVPR}, LRBP~\cite{Kong_Charless_2017_CVPR}, KP~\cite{Cui_2017_CVPR}, HIHCA~\cite{Cai_2017_ICCV} and SMSO~\cite{DBLP:conf/eccv/YuS18}) by a clear margin. Additionally, our MPN-COV-Net and its variants also outperform other deep GCP networks, i.e., B-CNN~\cite{lin2015bilinear} and improved B-CNN\cite{lin2017improved}. With ResNet-50 architecture, iSQRT-COV-Net (8K) respectively outperforms KP by about 2.6\%, 3.8\% and 0.6\% on Birds, Aircrafts and Cars, while iSQRT-COV-Net (32K) further improves accuracy. On all fine-grained datasets, existing methods employing 50-layer ResNet are no better than their counterparts with 16-layer VGG-VD. The reason may be that the last convolution layer of pre-trained ResNet-50 outputs 2048-dimensional features, much higher than 512-dimensional ones of VGG-VD, which are not suitable for existing second- or higher-order pooling methods. \textit{Different from all existing methods, our methods perform dimensionality reduction and deem second-order pooling as a component of CNN models, pre-trained on large-scale datasets.} Using pre-trained iSQRT-COV-Net with ResNet-101, we establish state-of-the-art results on three fine-grained benchmarks.

\subsubsection{Texture Classification}
We also transfer our iSQRT-COV-Net to texture classification, where DTD \cite{cimpoi14describing} and Indoor67 \cite{QuattoniT09} are employed. \emph{Indoor67} has 6,700 images from 67 indoor scene categories, where 80 and 20 images per-category are used for training and test, respectively.  \emph{DTD} consists of 5,640 material images collected from 47 classes, and pre-defined splits in \cite{cimpoi14describing} are used for evaluation. We adopt the same experimental settings with \cite{lin2015bilinear} for fair comparison. The results of different methods are listed in Table~\ref{table:texture}, where our iSQRT-COV-Net with VGG-VD16 architecture performs much better than the original model and clearly outperforms other deep second-order pooling networks, i.e., B-CNN \cite{lin2015bilinear} and FASON  \cite{Dai_2017_CVPR}. Meanwhile, our iSQRT-COV-Net is superior to Deep-TEN \cite{ZhangXD17} and MFAFV-Net \cite{DixitV16} (i.e., deep BoVW methods) using ResNet-50 and VGG-VD16 as backbone models, respectively. Additionally, iSQRT-COV-Net performs better than SMSO~\cite{DBLP:conf/eccv/YuS18} based on both VGG-VD16 and ResNet-50. The results on both FGVC and texture classification demonstrate our GCP methods can significantly improve the generalization (transfer) ability of deep CNNs.

\begin{table}[t]
	\footnotesize
	\renewcommand\arraystretch{1.15}
	\centering
	\caption{Classification accuracy ($\%$) of B-CNN and iSQRT-COV with different fine-tuning strategies on Birds (B), Aircrafts (A),  Cars (C), DTD (D) and Indoor67 (I) datasets.}
	\label{table:ft-com}
	\centering
		\begin{tabular}{c|c|c|c|c|c|c}
			\hline
			&  Method   & B   &  A  & C & D & I \\
			\hline
			\multirow{9}{*}{\rotatebox{90}{VGG-VD16}}   &  VGG-VD16 & 70.4  & 76.6      &  79.8      & 62.9$\pm$0.8  &  67.6  \\
			\cline{2-7}
			&  B-CNN & 84.0  & 83.9      &  90.6      & 72.9$\pm$0.8  &  79.0  \\
			\cline{2-7}
			& \tabincell{c}{B-CNN\\(PreTr-GCP)}   & 84.7  & 85.1      &  91.1        &  N/A &  N/A  \\
			\cline{2-7}
			& \tabincell{c}{iSQRT-COV\\(PreTr-GAP)}	  & 86.1  & 88.9      & 91.8        &  73.0$\pm$1.0 &  79.9  \\
			\cline{2-7}
			& \tabincell{c}{iSQRT-COV\\+ E-PN}        & 87.1  & 90.1      &  92.7     &  73.2$\pm$0.9  &  81.3  \\
			\cline{2-7}
			& iSQRT-COV-Net	       &               \textbf{87.2}  & \textbf{90.0}      &  \textbf{92.5}   &  \textbf{74.0$\pm$0.8}  &  \textbf{81.4} \\
			\hline
			\multirow{9}{*}{\rotatebox{90}{ResNet-50}} 	& ResNet-50~\cite{Cui_2017_CVPR}                                & 78.4    & 79.2      &  84.7  &  N/A &  N/A  \\
			\cline{2-7}
			&	B-CNN  & 84.4  & 85.4     &  91.4  &  N/A &  N/A  \\
			\cline{2-7}
			& 	\tabincell{c}{B-CNN\\(PreTr-GCP)}   & 85.2  & 86.1      &  91.5     &  N/A &  N/A  \\
			\cline{2-7}
			& \tabincell{c}{iSQRT-COV\\(PreTr-GAP)} &  86.7  & 89.6      &  92.2  & 73.4$\pm$0.9  &  82.5\\
			\cline{2-7}
			& \tabincell{c}{iSQRT-COV\\+ E-PN}      & 88.2  & 90.4    &  93.2    & 74.6$\pm$0.9   &  82.9\\
			\cline{2-7}
			& iSQRT-COV-Net	      &  \textbf{88.1}  & \textbf{90.0}  & \textbf{92.8}  & \textbf{74.8$\pm$1.0}   &  \textbf{83.5} \\  
			\hline
		\end{tabular}
	
\end{table}

\subsubsection{Effect of Different Fine-tuning Strategies.} Additionally, we evaluate how different fine-tuning strategies perform on small-scale datasets using  pre-trained models with GAP/GCP on large-scale ImageNet, and whether fine-tuning with E-PN~\cite{sanchez,lin2015bilinear} can further improve our proposed methods. Specifically, we employ standard pre-trained networks (i.e., networks with GAP) on ImageNet as backbone models, and replace GAP of the pre-trained models with our iSQRT-COV block while performing fine-tuning on fine-grained and texture benchmarks; for convenience, this  method is called iSQRT-COV (PreTr-GAP). As presented in Table~\ref{table:ft-com}, using both VGG-VD16 and ResNet-50 as backbone models, iSQRT-COV (PreTr-GAP) significantly improves simple fine-tuning of the original GAP-based ones on all benchmarks, indicating that the standard pre-trained networks also can greatly benefit from our GCP (i.e., iSQRT-COV). We also note that  iSQRT-COV-Net is superior to iSQRT-COV (PreTr-GAP), which indicates that pre-training the networks with GCP will bring improvement. It also  suggests that pre-training the exact models with a large-scale dataset performs better than the pre-training on a surrogate model. Furthermore, we train B-CNN with VGG-VD16 and ResNet-50 as backbone models on ImageNet, and the method is called B-CNN (PreTr-GCP); note that we insert a BN layer after $\ell_{2}$ normalization, otherwise it fails to converge. As compared in Table~\ref{table:ft-com}, iSQRT-COV clearly outperforms B-CNN under different settings, i.e., iSQRT-COV (PreTr-GAP) and iSQRT-COV-Net are superior to B-CNN and B-CNN (PreTr-GCP), respectively, showing high competitiveness of iSQRT-COV does not solely come from ImageNet pre-training of GCP networks. Finally, we pre-train the networks with iSQRT-COV and perform fine-tuning with E-PN, and  this method is indicated by iSQRT-COV + E-PN. The results of iSQRT-COV-Net \emph{vs.} iSQRT-COV + E-PN show that E-PN has little effect on our iSQRT-COV under the setting of fine-tuning.

\begin{table}[t]
	\footnotesize
	\centering
	\caption{Classification error  ($\%$) of different methods on iNaturalist Challenge 2018.}
	\label{table:iNaturalist}
	\centering
	\begin{tabular}{l|c|c}
		\hline
		Method & Description   &  Top-3 Err.  \\
		\hline
		\multirow{2}{*}{ResNet-152}	     & 320$\times$320 input        &  16.623   \\
		& 392$\times$392 input       &   16.024    \\
		\hline
		\multirow{2}{*}{iSQRT-COV-Net}	     & 320$\times$320 input       & 15.038   \\
		& 392$\times$392 input        &  14.704  \\
		\hline
		Our Fusion I        &  3 $\times$ ResNet-152 Model    & 14.625   \\
		\hline
		Our Fusion II       &  3 $\times$ iSQRT-COV-Net Model     & 13.409 \\
		\hline
		Our Fusion III       & Fusion I + Fusion II &\textbf{13.068}   \\
		\hline
		Runner-up &  Deep Learning Analytics & 14.214   \\
		\hline
	\end{tabular}
\end{table}

\subsection{iNaturalist Challenge 2018}\label{iNaturalist2018}
As part of the FGVC5 workshop at CVPR 2018, iNaturalist Challenge 2018 is a 
large-scale species classification competition, which contains over 8,000 species with 437,513 training, 24,426 validation and 149,394 test images. This dataset suffers from many visually similar species and high class imbalance, which are extremely difficult for accurate classification without expert knowledge.
We use \href{http://data.mxnet.io/models/imagenet-11k/resnet-152/}{ResNet-152 pre-trained on ImageNet-11K} as backbone model, and perform fine-tuning on training and validation sets. The top-3 error is reported on test server for comparison. To efficiently classify a large number of species, we use fast MPN-COV (i.e., iSQRT-COV with 3 iterations), and use a $1\times 1$ convolution with $160$ channels before GCP, leading to $\sim$12K dimensional representations, connected to 8142-way softmax classifier. Table~\ref{table:iNaturalist} gives the results of compared methods, where iSQRT-COV-Net distinctly outperforms the original ResNet-152 in configurations of both different input image sizes and ensemble models. By fusing iSQRT-COV-Net and ResNet-152 models, we positioned the first place, and outperforms the runner-up by ${\sim}1.2\%$. Note that only ensemble of iSQRT-COV-Net models is still better than the runner-up by ${\sim}0.8\%$, showing effectiveness of our  method.

\section{Conclusion}
This paper proposed a global Matrix Power Normalized COVariance (MPN-COV) Pooling methodology for improving the representation and generalization abilities of deep CNNs. Our matrix power normalization can not only estimate robustly covariance matrices but also can properly exploit Riemannian geometry of covariances. To further improve MPN-COV, we proposed a global Gaussian embedding method to integrate additional first-order statistical information, developed an iterative matrix square root normalization method for speeding up training of our covariance pooling networks, and studied a compact covariance representation strategy to reduce model complexity. As a result, our matrix power normalization methodology can improve significantly existing global covariance pooling methods, while enjoying fast training speed and affordable model complexity; meanwhile, it has solid statistical and geometrical foundations. The comprehensive experiments on both large-scale and small-scale benchmarks verified our MPN-COV and its improved solutions achieve a clear improvement with diversified deep CNN architectures, while obtaining state-of-the-art classification performance. In view of the effectiveness and architecture-independence of the proposed methods, they have the potential to be a commodity component in the existing deep CNN models, and they can motivate interests in future exploration of higher-order information in design of novel CNN architectures. In future, we will apply the proposed methods to other vision tasks, such as object detection \cite{RenHG017} and semantic segmentation \cite{ShelhamerLD17}.

\ifCLASSOPTIONcompsoc
  \section*{Acknowledgments}
\else
  \section*{Acknowledgment}
\fi

The work was supported by the National Natural Science Foundation of China (Grant No. 61971086, 61806140, U19A2073, 61471082, 61671182 and 61732011). Q. Wang was supported by National Postdoctoral Program for Innovative Talents. The work was done while Q. Wang 
was a PhD student at the School of Information and Communication Engineering, Dalian University of Technology, China.

\ifCLASSOPTIONcaptionsoff
  \newpage
\fi

\section*{Appendix I: Derivations for Backpropagation of iSQRT-COV Block}\label{Appendix1}

Though autograd toolkits provided by some deep learning frameworks can accomplish backpropagation of iSQRT-COV block automatically, their involved BP is still in a black box. Meanwhile, autograd toolkits sometimes bring uncertainty, e.g., autograd toolkit of PyTorch 0.3.0 or below cannot compute gradients of iSQRT-COV correctly. To make iSQRT-COV be self-contained and enable its implementation to be accessible when autograd toolkit is unavailable (e.g., the well-known~\href{http://caffe.berkeleyvision.org/}{Caffe} and early versions of~\href{http://www.vlfeat.org/matconvnet/}{MatConvNet}), we derive the gradients associated with the structured layers based on matrix backpropagation methodology~\cite{IonescuVS15}.  Below we take \textit{pre-normalization by trace} as an example, deriving the corresponding gradients.

\textbf{BP of Post-compensation} Given $\frac{\partial l}{\partial \mathbf{Z}}$ where $l$ is the loss function, the chain rule is of the form 
$
\mathrm{tr}\big(\big(\frac{\partial l}{\partial \mathbf{Z}}\big)^{T}\mathrm{d}\mathbf{Z}\big)=\mathrm{tr}\big(\big(\frac{\partial l}{\partial \mathbf{Y}_{N}}\big)^{T}\mathrm{d}\mathbf{Y}_{N}+\big(\frac{\partial l}{\partial \boldsymbol{\Sigma}}\big)^{T}\mathrm{d}\boldsymbol{\Sigma}\big) 
$. After some manipulations, we have
\begin{align}\label{equ:BP-post-trace}
\dfrac{\partial l}{\partial \mathbf{Y}_{N}}&=\sqrt{\mathrm{tr}(\boldsymbol{\Sigma)}}\dfrac{\partial l}{\partial \mathbf{Z}}  \nonumber \\
\dfrac{\partial l}{\partial \boldsymbol{\Sigma}}\Big|_{\mathrm{post}}&=\dfrac{1}{2\sqrt{\mathrm{tr}(\boldsymbol{\Sigma)}}}\mathrm{tr}\Big(\Big(\dfrac{\partial l}{\partial \mathbf{Z}}\Big)^{T}\mathbf{Y}_{N}\Big)\mathbf{I}.
\end{align}

\textbf{BP of Newton-Schulz Iteration} Then we  compute the partial derivatives with respect to $\frac{\partial l}{\partial \mathbf{Y}_{k}}$ and $\frac{\partial l}{\partial \mathbf{P}_{k}}$, $k=N-1, \ldots, 1$, given $\frac{\partial l}{\partial \mathbf{Y}_{N}}$ obtained by Eq.~(\ref{equ:BP-post-trace}) and  $\frac{\partial l}{\partial \mathbf{P}_{N}}=0$. As the matrix $\boldsymbol{\Sigma}$ is symmetric, it is easy to see from \textit{equation of Newton-Schulz iteration} (i.e., Eq.~(\ref{equ:coupled-equation})) that $\mathbf{Y}_{k}$ and $\mathbf{P}_{k}$ are both symmetric. According to the chain rules (omitted hereafter  for simplicity) of matrix backpropagation and after some manipulations, for $k=N, \ldots, 2$, we can derive
\begin{align}\label{equ:BP-coupled-equations}
\dfrac{\partial l}{\partial \mathbf{Y}_{k-1}}=&\dfrac{1}{2}\Big(\dfrac{\partial l}{\partial \mathbf{Y}_{k}}\Big(3\mathbf{I}-\mathbf{Y}_{k-1}\mathbf{P}_{k-1}\Big)-\mathbf{P}_{k-1}\dfrac{\partial l}{\partial \mathbf{P}_{k}}\mathbf{P}_{k-1} \nonumber\\  &-\mathbf{P}_{k-1}\mathbf{Y}_{k-1}\dfrac{\partial l}{\partial \mathbf{Y}_{k}}\Big)\nonumber\\
\dfrac{\partial l}{\partial \mathbf{P}_{k-1}}=&\dfrac{1}{2}\Big(\Big(3\mathbf{I}-\mathbf{Y}_{k-1}\mathbf{P}_{k-1}\Big)\dfrac{\partial l}{\partial \mathbf{P}_{k}}-\mathbf{Y}_{k-1}\dfrac{\partial l}{\partial \mathbf{Y}_{k}}\mathbf{Y}_{k-1} \nonumber\\ 
&-\dfrac{\partial l}{\partial \mathbf{P}_{k}}\mathbf{P}_{k-1}\mathbf{Y}_{k-1}\Big).
\end{align}
The final step of this layer is concerned with the partial derivative with respect to $\frac{\partial l}{\partial \mathbf{A}}$, which is given by
\begin{align}\label{equ:BP-gradient-A}
\dfrac{\partial l}{\partial \mathbf{A}}=\dfrac{1}{2}\Big(\dfrac{\partial l}{\partial \mathbf{Y}_{1}}\Big(3\mathbf{I}-\mathbf{A}\Big)-\dfrac{\partial l}{\partial \mathbf{P}_{1}}-\mathbf{A}\dfrac{\partial l}{\partial \mathbf{Y}_{1}}\Big).
\end{align}

\begin{table*}[h]
	\renewcommand\arraystretch{1.4}
	\caption{Comparison of different Gaussian embedding methods using VGG-VD16 on Birds-CUB200-2011 dataset. }
	\label{tab:GEF}
	\begin{center}
		\begin{tabular}{llc}
			\hline
			Method &  Gaussian Embedding & Accuracy $(\%)$\\
			\hline
			Nakayama et al. \cite{Nakayama-CVPR2010}  & $\mathbf{z}=\left[\begin{smallmatrix}
			vec(\boldsymbol{\Sigma}+\boldsymbol{\mu}\boldsymbol{\mu}^{T}), \boldsymbol{\mu}^{T}\end{smallmatrix}\right]^{T}$ & $83.5$ \\
			Calvo  et al. \cite{Calvo} or Lovri\'{c} et al.  \cite{RePEcjmvana} &
			$\left[\begin{smallmatrix}
			\boldsymbol{\Sigma}+\boldsymbol{\mu}\boldsymbol{\mu}^{T} & \boldsymbol{\mu}\\
			\boldsymbol{\mu}^{T} & 1
			\end{smallmatrix}\right]$ \vspace{2pt} &  $84.1$\\
			\cite{DBLP:conf/iccv/LiWZ13,DBLP:journals/tcsv/HuangWLLSGC18} (i.e., Calvo  et al. \cite{Calvo} or Lovri\'{c} et al. \cite{RePEcjmvana} + Log-Euclidean \cite{Arsigny2005})&
			$\log \left[\begin{smallmatrix}
			\boldsymbol{\Sigma}+\boldsymbol{\mu}\boldsymbol{\mu}^{T} & \boldsymbol{\mu}\\
			\boldsymbol{\mu}^{T} & 1
			\end{smallmatrix}\right]$ \vspace{2pt} & $83.8$\\
			\hline
			G$^2$DeNet (Ours) & $\left[\begin{smallmatrix}
			\boldsymbol{\Sigma}+\boldsymbol{\mu}\boldsymbol{\mu}^{T} & \boldsymbol{\mu}\\
			\boldsymbol{\mu}^{T} & 1
			\end{smallmatrix}\right]^{\frac{1}{2}}$ \vspace{2pt} & $\mathbf{87.1}$  \\
			\hline
		\end{tabular}
	\end{center}
\end{table*}

\begin{table*}[t]
	\centering
	\renewcommand\arraystretch{1.4}
	\caption{Comparison of our iSQRT-COV and GAP using various ResNets in terms of network parameters, floating point operations per second (FLOPs), inference time per image, memory cost for testing one image and classification error (\%, 1-crop prediction), given input images of size 224$\times$224.}\smallskip
	\begin{tabular}{lcccccc}
		\hline
		Methods & Parameter & GFLOPs. & Inference time (ms) & Memory Cost &Top-1 Err. & Top-5 Err.\\
		\hline
		ResNet18 + GAP & 11.69M & 1.82 & 0.60  & 763M & 30.24 & 10.92 \\
		ResNet18 + iSQRT-COV (32k)& 44.20M & 3.22 & 0.95 & 1263M & 24.52 & 7.77 \\
		ResNet18 + iSQRT-COV (8k) & 19.49M & 3.10 & 0.85 & 883M & 24.93 & 7.86 \\
		\hline
		ResNet34 + GAP & 21.80M & 3.67 & 0.88 & 901M & 26.70 & 8.58 \\
		ResNet34 + iSQRT-COV (32k)& 54.31M & 5.77 & 1.15 & 1407M & 22.89 & 6.64 \\
		ResNet34 + iSQRT-COV (8k) &29.71M & 5.56 & 1.10 & 1029M & 23.20 & 6.89 \\
		\hline
		ResNet50 + GAP & 25.56M & 3.86 & 1.29 & 975M & 24.70 & 7.80 \\
		ResNet50 + iSQRT-COV (32k) & 56.93M & 6.31 & 1.50 & 1463M & 22.14 & 6.22 \\
		ResNet50 + iSQRT-COV (8k) & 32.32M & 6.19 & 1.49 & 1085M & 22.33 & 6.28 \\
		\hline
		ResNet101 + GAP & 44.55M & 7.57 & 1.72 &  1233M & 23.60 & 7.10 \\
		ResNet101 + iSQRT-COV (32k) & 75.92M & 10.02 & 1.87 & 1733M & 21.21 & 5.68 \\
		ResNet101 + iSQRT-COV (8k) & 51.31M & 9.903 & 1.83  & 1355M & 21.32 & 5.70 \\
		\hline
		ResNet152 + GAP & 60.19M & 11.28 & 2.55 & 1483M & 23.00 & 6.70 \\
		\hline
	\end{tabular}
	\label{table:comparsion}	
\end{table*}

\textbf{BP of Pre-normalization} Note that here we need to combine the gradient of the loss function $l$ with respect to $\boldsymbol{\Sigma}$, backpropagated from the post-compensation layer. As such, by referring to equation of pre-normalization layer, we make similar derivations as before and obtain 
\begin{align}\label{equ:BP-pre-traceA}
\dfrac{\partial l}{\partial \boldsymbol{\Sigma}}=&-\dfrac{1}{({\mathrm{tr}(\boldsymbol{\Sigma)}})^2}\mathrm{tr}\Big(\Big(\dfrac{\partial l}{\partial \mathbf{A}}\Big)^{T}\boldsymbol{\Sigma}\Big)\mathbf{I}+\dfrac{1}{\mathrm{tr}(\boldsymbol{\boldsymbol{\Sigma}})}\dfrac{\partial l}{\partial \mathbf{A}} +\dfrac{\partial l}{\partial \boldsymbol{\Sigma}}\Big|_{\mathrm{post}}.
\end{align}

If we adopt \textit{pre-normalization by Frobenius norm}, the gradients associated with post-compensation  become
\begin{align}\label{equ:BP-post-fro}
\dfrac{\partial l}{\partial \mathbf{Y}_{N}}&=\sqrt{\|\boldsymbol{\Sigma}\|_{F}}\dfrac{\partial l}{\partial \mathbf{Z}} \nonumber \\
\dfrac{\partial l}{\partial \boldsymbol{\Sigma}}\Big|_{\mathrm{post}}&=\dfrac{1}{2\|\boldsymbol{\Sigma}\|_{F}^{3/2}}\mathrm{tr}\Big(\Big(\dfrac{\partial l}{\partial \mathbf{Z}}\Big)^{T}\mathbf{Y}_{N}\Big)\boldsymbol{\Sigma},
\end{align}
and that with respect to  pre-normalization  is
\begin{align}\label{equ:BP-pre-froA}
\dfrac{\partial l}{\partial \boldsymbol{\Sigma}}=&-\dfrac{1}{\|\boldsymbol{\Sigma}\|_{F}^{3}}\mathrm{tr}\Big(\Big(\dfrac{\partial l}{\partial \mathbf{A}}\Big)^{T}\boldsymbol{\Sigma}\Big)\boldsymbol{\Sigma}+\dfrac{1}{\|\boldsymbol{\Sigma}\|_{F}}\dfrac{\partial l}{\partial \mathbf{A}} +\dfrac{\partial l}{\partial \boldsymbol{\Sigma}}\Big|_{\mathrm{post}},
\end{align}
while the backward gradients of Newton-Schulz iteration (\ref{equ:BP-coupled-equations}) keep unchanged. Based on above derivations, we can achieve the backpropagation of iSQRT-COV block without autograd toolkit.  

\section*{Appendix II: Comparison of Different Gaussian Embedding Methods for G$^2$DeNet}\label{Appendix2}

To insert a global Gaussian distribution into deep CNNs, we introduce a Gaussian embedding method studied  in~\cite{L2EMG} to identify a Gaussian $\mathcal{N}(\boldsymbol{\mu},\boldsymbol{\Sigma})$ as a square root SPD matrix (i.e., $\left[\begin{smallmatrix} \boldsymbol{\Sigma}+\boldsymbol{\mu}\boldsymbol{\mu}^{T} & \boldsymbol{\mu}\\ \boldsymbol{\mu}^{T} & 1 \end{smallmatrix}\right]^{\frac{1}{2}}$). Although many Gaussian embedding methods have been studied, there exist clear differences between ours and them. Specifically, Nakayama et al. \cite{Nakayama-CVPR2010} embed Gaussians in a flat manifold by taking an affine coordinate system. In \cite{GongWang2009}, Gaussian is mapped to a unique positive definite lower triangular affine transform (PDLTAT) matrix (i.e., $\left[\begin{matrix} \mathbf{L} & \boldsymbol{\mu} \\ \mathbf{0}^{T} & 1 \end{matrix}\right]$ and $\mathbf{L}\mathbf{L}^{T}$ is the Cholesky decomposition of $\boldsymbol{\Sigma}$), whose space forms an affine group. Such Gaussian embedding is not suitable for backpropagation due to involvement of Cholesky decomposition of covariance matrix $\boldsymbol{\Sigma}$. The methods in Calvo et al. \cite{Calvo} and Lovri'c et al. \cite{RePEcjmvana} respectively embed the space of Gaussian in the Siegel group and the Riemannian symmetric space, identifying a Gaussian as a unique SPD matrix, i.e., $\begin{bmatrix}
\boldsymbol{\Sigma}+\boldsymbol{\mu}\boldsymbol{\mu}^{T} & \boldsymbol{\mu}\\ \boldsymbol{\mu}^{T} & 1
\end{bmatrix}$ or $C\ast\begin{bmatrix}
\boldsymbol{\Sigma}+\boldsymbol{\mu}\boldsymbol{\mu}^{T} & \boldsymbol{\mu}\\ \boldsymbol{\mu}^{T} & 1
\end{bmatrix}$ where $C$ is a scalar. After that, \cite{DBLP:conf/iccv/LiWZ13,DBLP:journals/tcsv/HuangWLLSGC18} employ Log-Euclidean Riemannian metric (LERM) \cite{Arsigny2005} on the resulting embedding matrix in \cite{Calvo,RePEcjmvana}, which can be presented as $\log\bigg(\begin{bmatrix}
\boldsymbol{\Sigma}+\boldsymbol{\mu}\boldsymbol{\mu}^{T} & \boldsymbol{\mu}\\ \boldsymbol{\mu}^{T} & 1
\end{bmatrix}\bigg)$. Different from these methods, our introduced embedding method considers both geometric and algebraic (i.e., Lie group) structures of Gaussian, while achieving better performances. In particular, we compare our G$^2$DeNet with various Gaussian embedding methods on Birds-CUB200-2011 dataset using VGG-VG16 as backbone model. The results are given in Table~\ref{tab:GEF}, from it we can see that our introduced embedding method achieves the best performance, outperforming other competing methods by $3\%\sim3.6\%$.

\section*{Appendix III: Comparison of iSQRT-COV and GAP in terms of efficiency and effectiveness using ResNets}\label{Appendix3}

In this section, we compare global average pooling (GAP) and our proposed global covariance pooling (GCP) on large-scale ImageNet using ResNet-18, ResNet-34, ResNet-50 and ResNet-101 as backbone models. The evaluation metrics include network parameters, floating point operations per second (FLOPs), inference (FP computation) time per image, memory cost for testing one image and Top-1/Top-5 classification errors, given input images of size 224$\times$224. For our GCP, we use the iSQRT-COV-Net by setting covariance representations to 32k and 8k, respectively. All models are trained with the same experimental settings and run on a workstation equipped with four Titan Xp GPUs, two Intel(R) Xeon Silver 4112 CPUs @ 2.60GHz, 64G RAM and 480 GB INTEL SSD.  From the comparison results in Table~\ref{table:comparsion}, we have three observations as follows. (1) Both our iSQRT-COV (32k) and iSQRT-COV (8k) significantly outperform the original first-order GAP under the same CNN architectures. In particular, iSQRT-COV (8k) achieves clear gains with affordable computational cost. Specifically, it introduces extra $\sim$7M parameters, $\sim$0.2ms inference time and $\sim$100M memory cost, but decreases about 5.3\%, 3.5\%, 2.4\% and 2.3\% Top-1 errors over the original GAP-based ResNet-18, ResNet-34, ResNet-50 and ResNet-101, respectively. (2) When we compare different models that have matching performance, we can see that  iSQRT-COV (8k) models  show  much lower  complexity than the corresponding GAP models. For example, ResNet34 + iSQRT-COV (8k) and ResNet50 + iSQRT-COV (8k) are comparable or slightly superior to ResNet101 + GAP and ResNet152 + GAP in performance, respectively, whereas the former ones have much lower complexity than the latter ones. (3) When different models that have similar space and time complexity are compared, it can be seen that iSQRT-COV (8k) models clearly outperform the corresponding GAP models. For example, ResNet18 + iSQRT-COV (8k), ResNet50 + iSQRT-COV (8k) and ResNet101 + iSQRT-COV (8k) have similar (or lower)  complexity with ResNet34 + GAP, ResNet101 + GAP and ResNet152 + GAP, but drop 1.8\%, 1.3\% and 1.7\% in terms of Top-1 error, respectively. Above results demonstrate that our iSQRT-COV has better ability to balance efficiency and effectiveness.

\begin{table}[htb]
	\renewcommand\arraystretch{1.4}
	\caption{Classification accuracy (\%) of  iSQRT-COV-Net with the softmax and SVM classifiers on fine-grained visual classification task.}
	\label{table:softmax}
	\centering
	\begin{tabular}{lccccc}
		\hline
		& Backbone	& Classifier  & Birds   &  Aircrafts  & Cars  \\
		\hline 
		\multirow{4}{*}{\tabincell{c}{iSQRT-COV\\-Net}}	& \multirow{2}{*}{VGG-D16}
		& SVM               &               87.2  & 90.0      &  92.5\\
		&	& softmax               &               87.1  & 90.0      &  92.5\\
		\cline{2-6}
		&	\multirow{2}{*}{ResNet-50}
		& SVM & 88.1  & 90.0  & 92.8 \\
		&	& softmax               &               88.1  & 90.5      &  92.7\\
		\hline
	\end{tabular}
\end{table}

\section*{Appendix IV: Comparison between SVM and Softmax classifiers on Fine-grained Visual Classification Task}\label{Appendix4}
For the purpose of fair comparison with existing works, we in fact adopt exactly the same experimental settings with~\cite{lin2015bilinear,Gao_2016_CVPR,Cui_2017_CVPR,Cai_2017_ICCV}, where an SVM is trained for classification after fine-tuning models. To assess the effect of classifiers on our method, we provide the results with softmax classifier used at training time, as presented in Table~\ref{table:softmax}. We can see that SVM and softmax classifiers achieve very similar results for our iSQRT-COV-Net, except iSQRT-COV-Net with ResNet-50  on Aircrafts, where softmax classifier is a little better than SVM classifier. The experiments suggest that, for our iSQRT-COV-Net, overall training SVM classifier after fine-tuning with softmax classifier brings negligible benefits.

\begin{table*}[thb]
	\renewcommand\arraystretch{1.4}
	\caption{Comparison of iSQRT-COV and Improved B-CNN~\cite{lin2017improved} in terms of memory cost, training speed and classification accuracy (\%). Following the  experimental settings of Improved B-CNN~\cite{lin2017improved}, we employ VGG-VD16 as backbone model and set batch-size to 24, while performing fine-tuning on Birds-CUB200-2011 dataset. Here, we give memory taken by GCP layer for the whole mini-batch, percentage of memory overhead due to caching of intermediate tensors ($+\Delta$) with respect to that of the whole networks (ALL), training speed (Hz) and classification accuracy. T indicates number of iterations.}
	\label{table:memory}
	\centering
	\begin{tabular}{c|l|l|c|c|c|l}
		\hline
		Methods &  \parbox{22mm}{\centering FP} & \parbox{20mm}{\centering BP} & \parbox{20mm}{\centering{Memory for \\\textbf{GCP layer}}} & $+\Delta$/ALL &\parbox{20mm}{\centering{Training speed \\ (Hz)}} & Accuracy\\
		\hline
		\multirow{4}{*}{Improved B-CNN~\cite{lin2017improved}} & NS iteration (T=5) & SVD+lyap & \multirow{4}{*}{224M + 0M} & 0   & 17.13 & 85.8~\cite{lin2017improved}\\
		& NS iteration (T=5) & iterative Lyap solver (T=5) &  & 0   & 29.41  & 84.8\\
		& NS iteration (T=10) & iterative Lyap solver (T=10) &  & 0   & 28.30 & 85.2\\	
		& NS iteration (T=20) & iterative Lyap solver (T=20) &  & 0   & 26.12 & 85.7\\	
		\hline
		iSQRT-COV   & NS iteration (T=5) & NS iteration (T=5) & 224M + 96.18M & 0.89\% & 31.25 & 87.2 \\
		\hline
	\end{tabular}
\end{table*}

\section*{Appendix V: Comparison of iSQRT-COV and Improved B-CNN in terms of Memory, Speed and Accuracy}\label{Appendix5}

Here, we compare iSQRT-COV with Improved B-CNN~\cite{lin2017improved} in term of memory cost, training speed and classification accuracy. Our induces memory overhead as the intermediate  tensors in the Newton-Schulz iteration  have to be cached for usage in the backpropagation; by contrast, the methods proposed in improved B-CNN~\cite{lin2017improved} do not involve this memory overhead. We conduct additional experiments to evaluate speed and memory cost of different methods. The improved B-CNN proposed several different fine-tuning methods, including FP and BP both dependent upon SVD, FP by Denman-Beavers iteration or Newton-Schulz (NS) iteration while BP by SVD$+$Lyapunov (Lyap) equation. We compare with the  method that uses NS iteration for FP and SVD$+$Lyap for BP, which is more efficient among all the proposed methods and allow accurate gradients computation. The authors implement, in their released code, a new method that was not described in their paper, in which gradients can be computed by iterative Lyap solver, avoiding time consuming SVD; we also compare with this method.  We make experiments on Birds-CUB200-2011 dataset following the  experimental settings of improved B-CNN~\cite{lin2017improved}. We employ VGG-VD16 as backbone model and set batch-size to 24. To implement improved B-CNN, we use the source code released by the authors. We employ the function 'sqrt\_newton\_schulz' for FP by NS iteration, the function 'sqrt\_svd\_lyap'  for BP by SVD$+$Lyap, and the function  'lyap\_newton\_schulz' for BP by  iterative Lyap solver. The experiments  run on a PC equipped with a 4-core Intel i7-4790k@4.0GHz CPU, 32G RAM, 512GB Samsung PRO SSD and a single GTX 1080Ti. The results are presented in Table~\ref{table:memory}, where iSQRT-COV brings extra memory (i.e., $+\Delta$) as it needs to cache intermediate tensors. Nevertheless, the percentage of memory overhead with respect to that of the whole networks (ALL) is negligible ($<$1\%). The training speed of iSQRT-COV is 1.8x faster than improved B-CNN based on NS iteration and SVD+lyap. Note that, as shown in project page of the authors, iterative Lyapunov solver is an approximation algorithm, leading to larger error. Therefore, it requires more iterations for achieving satisfying results. When number of iterations is 5, improved B-CNN based on NS iteration and Lyap iteration has comparable training speed with iSQRT-COV, but iSQRT-COV has higher classification accuracy. More iterations will bring some improvement for improved B-CNN based on NS iteration and Lyap iteration, but it slows down training speed. Overall, iSQRT-COV can make a good balance for memory cost, training speed and classification accuracy.

\bibliographystyle{IEEEtran}
\bibliography{IEEEabrv,egbib}

\end{document}